\documentclass[12pt]{article}

\usepackage[utf8]{inputenc}
\usepackage[margin=1in]{geometry}
\usepackage{graphicx}
\usepackage{amsmath}
\usepackage{amssymb}
\usepackage{booktabs}
\usepackage{hyperref}
\usepackage{xcolor}
\usepackage{enumitem}
\usepackage{longtable}
\usepackage{multirow}
\usepackage[toc,page]{appendix}
\usepackage[breakable, skins, listings]{tcolorbox}
\usepackage{amssymb}
\usepackage{enumitem}
\usepackage{subcaption}
\usepackage{rotating}

\title{A Real-World Evaluation of LLM Medication Safety Reviews in NHS Primary Care}

\author{%
  \parbox{\textwidth}{\centering \small
    Oliver Normand\textsuperscript{1},
    Esther Borsi\textsuperscript{1},
    Mitch Fruin\textsuperscript{1},
    Lauren E Walker\textsuperscript{2,3},
    Jamie Heagerty\textsuperscript{4},
    Chris C. Holmes\textsuperscript{5,6},
    Anthony J Avery\textsuperscript{7},
    Iain E Buchan\textsuperscript{3},
    Harry Coppock\textsuperscript{1,4,8,9}\\[0.75em]
    \textsuperscript{1}i.AI, Department for Science, Innovation, and Technology\\
    \textsuperscript{2}Centre for Experimental Therapeutics, University of Liverpool\\
    \textsuperscript{3} Civic Health Innovation Labs, University of Liverpool\\
    \textsuperscript{4}10 Downing Street\\
    \textsuperscript{5}Department of Statistics, University of Oxford\\
    \textsuperscript{6}Ellison Institute of Technology\\
    \textsuperscript{7}Centre for Academic Primary Care, University of Nottingham\\
    \textsuperscript{8}The UK AI Security Institute\\
    \textsuperscript{9}Department of Computing, Imperial College London\\
  }%
}

\date{\today}

\begin{document}

\maketitle 
\vspace{-1cm}
\begin{abstract}
    Large language models (LLMs) often match or exceed clinician-level performance on medical benchmarks, yet very few are evaluated on real clinical data or examined beyond headline metrics. We present, to our knowledge, the first evaluation of an LLM-based medication safety review system on real NHS primary care data, with detailed characterisation of key failure behaviours across varying levels of clinical complexity. In a retrospective study using a population-scale EHR spanning 2,125,549 adults in NHS Cheshire and Merseyside, we strategically sampled patients to capture a broad range of clinical complexity and medication safety risk, yielding 277 patients after data-quality exclusions. An expert clinician reviewed these patients and graded system-identified issues and proposed interventions. Our primary LLM system showed strong performance in recognising when a clinical issue is present (sensitivity 100\% [95\% CI 98.2--100], specificity 83.1\% [95\% CI 72.7--90.1]), yet correctly identified all issues and interventions in only 46.9\% [95\% CI 41.1--52.8] of patients. Failure analysis reveals that, in this setting, the dominant failure mechanism is contextual reasoning rather than missing medication knowledge, with five primary patterns: overconfidence in uncertainty, applying standard guidelines without adjusting for individual patient context, misunderstanding how healthcare is delivered in practice, factual errors, and process blindness. These patterns persisted across patient complexity and demographic strata, and across a range of state-of-the-art models and configurations. We provide 45 detailed vignettes that comprehensively cover all identified failure cases, including patient context, LLM output, and expert clinical judgements. This work highlights shortcomings that must be addressed before LLM-based clinical AI can be safely deployed. It also begs larger-scale, prospective evaluations and much deeper study of LLM behaviours in clinical contexts.

\end{abstract}
\tableofcontents 
\section{Introduction}
    \label{sec:Introduction}
    
    Large Language Models (LLMs) are rapidly surpassing human expert performance across many domains\cite{OpenAI2024O4MiniSystemCard,Anthropic2024SystemCard,DeepSeekAI2025R1, Kwa2025Measuring, epoch-benchmark, LuongLockhart2025} with no signs of slowing down~\cite{Arora2025HealthBench, Tu2025ConversationalAI, aisi_oneyear}. Healthcare is no exception: recent benchmarks and evaluations demonstrate that state-of-the-art language models now achieve or exceed clinician-level accuracy on a wide range of medical tasks~\cite{Arora2025HealthBench,Tu2025AMIE, Korom2025, bedi2025medhelmholisticevaluationlarge, Johnson2019MIMICCXR, deyoung2021ms2multidocumentsummarizationmedical, jin2019pubmedqadatasetbiomedicalresearch, jin2020whatdisease, sellergren2025medgemmatechnicalreport}. 
    
    \subsection{The Medication Crisis}
        \label{sec:med_crisis}

        Medicines are central to modern health care, but safe prescribing is increasingly challenging in an ageing population living with multiple long-term conditions. The World Health Organization (WHO) estimates that medication errors cost approximately \$42\,billion per year globally (almost 1\% of total global health expenditure)~\cite{WHO_MedicationWithoutHarm,WHO_2017_MedicationErrorCost}. Preventable medication-related harm affects around 5\% of patients, with more than a quarter of events being severe or life-threatening~\cite{Panagioti2024GlobalBurdenPreventableMedHarm}. Medication-related issues account for an estimated half of avoidable medical harm~\cite{Hodkinson2020PreventableMedicationHarm}.

        In the UK, over 1.2 billion prescriptions are issued annually at a cost of £17 billion~\cite{LSE2023CostsNewDrugs,NHSDigital2020PrescribingCosts}. Medication errors cost the NHS between £98.5 million and £1.6 billion annually, cause between 1,700 and 22,000 deaths each year, and account for an estimated 8\% of hospital admissions\cite{BMJ2020MedicationErrors,Osanlou2022ADRMultimorbidityPolypharmacy}. While targeted medication reviews significantly reduce adverse events~\cite{Avery2012, Odeh2020MedicinesOptimisation}, the NHS currently lacks the capacity to meet review targets, resulting in reactive rather than preventative care~\cite{NHSDigital2024GPContract}.
    
\subsection{Progress in Clinician-Level Artificial Intelligence}
\label{sec:progess_llm}

Recent studies show substantial progress toward clinician-level medical reasoning in large language models (LLMs). In interactive clinical settings, AMIE outperformed primary care physicians on 30 of 32 diagnostic and communication metrics in simulated patient interviews, and it also improved clinician diagnostic performance when used as a decision-support tool on published case reports~\cite{Tu2025AMIE,McDuff2025AMIE}. Scaling medical foundation models has yielded similar gains: MedFound, a 176B-parameter model, outperformed strong baselines on curated clinical benchmarks and retrospective real patient records~\cite{Liu2025MedFound}. Evidence of clinical impact is also emerging; in a randomised controlled trial, GPT-4 assistance reduced the risk of harm in physician decision-making on complex cases~\cite{Chen2025GPT4}. Finally, large-scale physician-graded evaluations suggest rapid capability improvements over successive model generations: on HealthBench (5,000 realistic health conversations graded by 262 physicians), o3 scored 60\% compared to GPT-4o's 32\%~\cite{Arora2025HealthBench}.

    \subsection{The Evaluation Gap}
        \label{sec:evaluation-gap}

        Despite rapid LLM adoption in healthcare~\cite{elsevier_clinician_future_2025}, recent systematic reviews reveal critical gaps in current research practices and coverage. In particular, only 5\% of studies use real patient data; the vast majority relying on synthetic patient data or medical exam style questions~\cite{suhana2025}. This reliance on examination-style evaluation is particularly problematic as high performance on knowledge-based benchmarks doesn't translate to equivalent clinical competence~\cite{gong2025knowledgepracticegap}, and even expert-level accuracy can mask substantial reasoning flaws~\cite{Jin2024HiddenFlaws}. Only a small fraction of these studies reported on failure mechanisms and most relied on automated scoring mechanisms which are known to miss clinically salient hallucinations and omissions~\cite{asgari_framework_2025}. We focus on the few recent works that both evaluate LLMs in real clinical settings and attempt to characterise their failure modes.
        
        \paragraph{Clinical Settings} 
            Several studies have begun evaluating LLMs in realistic clinical scenarios, including real patient records and open-ended decision-making tasks. For example, MedFound~\cite{Liu2025MedFound} was tested on retrospective clinical records spanning eight specialities, and Gaber et al.~\cite{gaber_evaluating_2025} benchmark Claude-family models and a Retrieval-Augmented Generation (RAG)-assisted workflow on 2{,}000 anonymised public patient Electronic Health Records (EHR) cases for triage, referral, and diagnosis, finding moderate gains from RAG and consistent but imperfect performance across user types (patient-style vs clinician-style inputs).
        
            Complementing these retrospective evaluations, Korom et al~\cite{korom2025aibasedclinicaldecisionsupport}. study a GPT-4o–based “safety net”, integrated into the EHR across 15 high-volume primary care clinics in Nairobi. In a clinical trial covering 39{,}849 visits, clinicians with access to AI made 32\% fewer history-taking errors, 10\% fewer investigation errors, 16\% fewer diagnostic errors, and 13\% fewer treatment errors than colleagues without access, with no detected instances where AI advice caused harm and only modest, non-significant differences in short-term patient-reported outcomes. 
        
            These studies provide early evidence that LLMs can reduce clinical errors when embedded into workflows. However, they focus on narrow outcome sets—primarily diagnostic accuracy and aggregated error rates in a single health system—and treat model behaviour largely as a black box, offering limited insight into how failures manifest across the full clinical encounter (for example, through subtle hallucinations, omissions, or brittle reasoning). 
        
        \paragraph{Failure Analysis} 
            A small but growing body of work has examined the brittleness of medical reasoning in LLMs. Bedi et al.~\cite{bedi2025} modified USMLE-style questions so that the correct answer was ``None of the above.''; state-of-the-art models showed dramatic performance drops (e.g. 85\% to 60\%), suggesting apparent competence relies partly on pattern recognition rather than robust reasoning. Models also exhibit systematic metacognitive deficiencies, failing to recognise knowledge limitations and providing confident answers even when correct options are absent~\cite{Griot2025}. Kim et al.~\cite{kim2025medicalhallucinationsfoundationmodels} evaluated 11 foundation models across seven medical benchmarks and found that most residual errors arose from causal and temporal reasoning failures rather than knowledge gaps. Asgari et al.~\cite{asgari_framework_2025} proposed CREOLA, a clinician-in-the-loop framework for assessing hallucinations and omissions in LLM-generated consultation notes, finding that hallucinations were far more likely than omissions to be clinically significant.

        \paragraph{Medication Safety}
            A recent systematic review of AI use in medication safety identified only one study that evaluated systems in real clinical settings; the vast majority of work relied on synthetic vignettes or multiple-choice questions. Across studies, evaluation focused predominantly on technical performance metrics (e.g.\ accuracy, F1), with the clinical consequences of failures rarely reported. The authors concluded that, despite significant attention in the literature, it remains unclear whether AI systems reduce medication-related harm in practice or what the consequences of deployment would be for patient outcomes~\cite{ong2025scoping}. \\
    
        \paragraph{This Study}
            This evaluation gap in setting diversity, task coverage, and failure-mode characterisation motivates this study. Here, we evaluate a range of state-of-the-art open source models in real clinical settings, and present a framework on failure mechanisms observed with example vignettes. This study probes not just \textit{whether} LLMs work, but \textit{how} and \textit{where} they fail in clinical contexts. We identify five distinct failure reasons that account for 178 failures across 148 patients, revealing that contextual reasoning failures outnumber factual errors 6:1. We hope this study informs understanding of behaviours in these real world settings and motivates further work and full clinical trials so the benefits of this technology can be fully realised by the public.

\section{Methods}
\subsection{Study Design and Setting}

    This study was a cross-sectional evaluation using electronic health records (EHR) data from the NHS Cheshire and Merseyside Integrated Care Board. It was conducted within a Trusted Research Environment (TRE) provided by GraphNet Ltd. The study received approval from the Data Access and Asset Group following completion of data protection impact assessment, with access granted from June 8th to November 8th 2025.

\subsection{Data Source and Population}
    The dataset comprised 2,125,549 unique adults aged 18 and over with linked GP events, medications, hospital episodes, and coded observations using SNOMED CT and dm+d coding systems. From this EHR data, we constructed structured \textit{patient profiles} --- longitudinal records containing all available clinical information for each patient. These patient profiles contained only coded information; no free-text clinical notes were available in the EHR export.

    Among patients with at least one recorded GP event (98.8\%), the mean number of events per patient was 1,010 (median 666, range 1-31,438, SD 1,126). Dates ranged from 1976 to 2025, with the interquartile range spanning 2012 to 2022. The population was more deprived than the national average: the Index of Multiple Deprivation (IMD) distribution showed 26.6\% of patients in England's most deprived decile compared with 7.7\% in the least deprived decile.

\subsection{System Architecture}
    \label{sec:system-architecture}
    We evaluated \texttt{gpt-oss-120b}, a 120-billion parameter model released in August 2025 by OpenAI, using the Inspect evaluation framework~\cite{ukaisecurity2024inspectai}. This LLM-based medication safety review configuration is referred to throughout as ``the System". The system prompt specified the task (identify medication safety issues requiring intervention), described the input format, and defined the expected output structure. It avoided specific methodological frameworks or examples. The complete system prompt is provided in Appendix~\ref{app:system-prompt}. The System operated without external knowledge sources such as  the BNF, NICE Guidelines, or PubMed. Each patient profile was converted to chronologically ordered markdown. The System generated structured outputs including binary intervention flag, probability score, identified issues with evidence, and proposed interventions.

\subsection{Case Sampling Strategy}

    The evaluation considered patients across a spectrum of clinical complexity, defined by age, comorbidity burden (count QoF registers belonged to), number of active medications, and frequency of recent GP events. From a 200,000 patient test set, we selected 300 patients for expert review. Three sampling strategies were employed to achieve this coverage.

    Firstly, prescribing safety indicators (detailed in Appendix \ref{app:prescribing-safety-indicators}) were used to identify high-complexity cases with medication safety concerns. From the 8,326 patients meeting at least one indicator in the 200,000 patient test set, 100 patients were randomly sampled with stratification across all 10 indicators to ensure diverse representation.

    In addition, 100 indicator-negative patients were selected using stratified matching on age, sex, number of active prescriptions, and GP events since 2020. These cases exhibited similar baseline clinical complexity to positive cases.

    To better understand System performance in less complex clinical scenarios and enable extrapolation to population level performance, an additional 100 patients were sampled from the indicator-negative population. The System was first applied to a larger indicator-negative sample, then 50 System-positive cases (flagged for intervention) and 50 System-negative cases (no intervention flagged) were randomly selected for clinician review.

    This combined sampling approach yielded 300 patients. After excluding 23 cases with data quality issues, 277 patients remained for analysis. 
    
    Given the role of prescribing safety indicators in case sampling, we separately analysed System performance against these criteria and clinician-indicator agreement rates; results are provided in Appendix~\ref{app:prescribing-safety-indicators}. Likewise, the subset of randomly sampled indicator-negative patients enables extrapolation to population-level performance estimates; results are provided in Appendix~\ref{app:population-performance-assessment}.

\subsection{Clinician evaluation of patients}

    All analysed cases underwent independent review by an experienced clinician with expertise in medication optimisation and safety. This clinician-evaluated design followed preliminary analysis demonstrating that ground truth cannot be reliably established from coded EHR data alone: medication change rates were nearly identical between structured medication reviews coded as identifying issues versus those coded as finding none (30.8\% vs 30.2\% within three months).
    
    The clinician was tasked purely with System evaluation and had no part in System development. They were provided access to patient profiles via a custom-built evaluation interface hosted locally within the TRE, which displayed the same information as the System. The interface displayed patient profiles chronologically and allowed systematic assessment across multiple dimensions. Examples of the evaluation interface are provided in Appendix~\ref{app:evaluation-app}. For each patient the clinician evaluated:

    \begin{enumerate}[nosep]
        \item For each issue identified by the System, whether this was correct or not
        \item Whether the System missed any other issues
        \item For the intervention proposed by the System, whether this was correct or not
    \end{enumerate}
    
    \noindent
    The test cases were reviewed between October and November 2025.

\subsection{Three-Level Hierarchical Evaluation Framework}

    Binary classification (patient has an issue: yes/no) provides insufficient granularity to understand where and why medication safety systems fail. We use a hierarchical evaluation to record System performance at three levels, all marked by the clinician.

    Similar multi-level frameworks have recently been proposed for diagnostic reasoning evaluation, where performance degradation across stages reveals reasoning brittleness invisible to single-metric assessment~\cite{Qiu2025}. 

    \textbf{Level 1: Issue Identification.} Did the System correctly flag any issue in the patient where one was present? This is marked by the clinician during their review.

    \textbf{Level 2: Issue Correctness (conditional on Level 1).} Among cases where the System flagged an issue, did it identify the correct issues? Cases failing Level 1 (not flagged) do not progress to Level 2 evaluation.

    \textbf{Level 3: Intervention Appropriateness (conditional on Level 2).} Among cases where the System correctly identified all the issues, did it propose a correct intervention? Cases failing Level 2 (wrong issue identified) do not progress to Level 3 evaluation.

\subsection{Failure Mode Analysis}
    To characterise System failure modes, we reviewed all System analyses exhibiting errors at any hierarchical level. Individual cases could exhibit multiple failure patterns. Themes were developed iteratively. Through this process, a natural distinction emerged between \textit{failure reasons}---the underlying causes explaining why the System failed---and \textit{failure modes}---the specific observable manifestations of how failures presented.

    Where a single patient exhibited multiple distinct failure patterns, each was coded separately, yielding 178 failure instances from 148 patients. The resulting five-category taxonomy (Figure~\ref{fig:figure2}) extends prior frameworks for medical AI failure analysis~\cite{kim2025medicalhallucinationsfoundationmodels, asgari_framework_2025} to the medication safety domain. Representative vignettes illustrating each failure mode are provided in Appendix~\ref{app:failure-vignettes}.

    To assess the potential clinical significance of System failures, each failure instance was independently classified by an experienced clinician according to the WHO International Classification for Patient Safety harm categories~\cite{Cooper2018WHOClassification}, adapted to assess likely outcome if the System's recommendation were implemented without further review. Categories ranged from none (no symptoms, no treatment required) through mild, moderate, and severe to death.

\subsection{Performance by Patient Complexity}
    We assessed performance stratified by patient age, number of active medications, and QoF register membership count (proxy for co-morbidity burden). Given potential confounding between these measures, we assessed their intercorrelation using Pearson coefficients and performed multiple regression to identify independent predictors of performance.

    The clinician did not provide an overall score for each patient. Instead, a composite score ($S_{\text{clinician}}$, range 0--1) was calculated from individual assessments, combining issue identification (as an $F_1$ score) and intervention appropriateness (full/partial/no credit). The scoring methodology assigns partial credit (0.5) for partially correct outputs, reflecting clinical judgment that these retain some utility while being clearly distinguished from fully correct assessments. Full specification is provided in Appendix~\ref{app:scoring}.

\subsection{Automated Scoring}

    The methods so far relied on clinician evaluation of all 277 cases. Further insight relied on quantitative metrics across multiple models, complexity strata and repeated runs, which is not feasible manually.

    We developed an automated scorer using the clinician's feedback on the 277 patients as ground truth. Clinician assessments were synthesised by \texttt{gpt-oss-120b} into lists of clinical issues and interventions. A separate LLM judge then assessed alignment between System outputs and this ground truth, producing an automated score ($S_{\text{automated}}$, range 0--1) combining $F_1$ scores for issue identification and intervention appropriateness. True negatives (both system and ground truth agree no issues exist) score 1.0; disagreements on issue presence score 0.0. Full specification is provided in Appendix~\ref{app:scoring}.

    $S_{\text{automated}}$ was used to assess model self-consistency, compare performance across models, and evaluate variation with stated patient ethnicity. Alternative scoring methods that weight false negatives more heavily than false positives could better reflect the asymmetric harm from missing genuine risks in any deployment.

    \subsubsection{Self-Consistency and Anchoring Bias Assessment}
    \label{sec:self-consistency}
        To quantify anchoring bias in clinician evaluation, we calculated the System's inherent accuracy ceiling from self-consistency data. From 10 repeated runs per patient, we measured: $P(\text{re-flag } | \text{ initially flagged})$ and $P(\text{maintain negative } | \text{ initially negative})$. Applying these probabilities to the 277 evaluable patients yields expected true positives and true negatives based solely on model behaviour. The difference between this model self-consistency ceiling and observed clinician-agreement accuracy (95.7\%) represents the minimum anchoring effect. 

    \subsubsection{Multi-Model Comparison}
        To assess whether performance patterns were model-specific or generalisable, the System was evaluated using six model configurations spanning four distinct foundation models: \href{https://huggingface.co/openai/gpt-oss-120b}{\texttt{gpt-oss-120b}} with three reasoning effort configurations (low, medium, high), \href{https://huggingface.co/openai/gpt-oss-20b}{\texttt{gpt-oss-20b}} (medium reasoning), \href{https://huggingface.co/google/gemma-3-27b-it}{\texttt{gemma-3-27b-it}}, and \href{https://huggingface.co/google/medgemma-27b-text-it}{\texttt{medgemma-27b-text-it}}. Inspect enabled identical evaluation pipelines across all models without code modification~\cite{ukaisecurity2024inspectai}.

        Each model was evaluated with 10 independent repeats over the test set to characterise both mean performance and run-to-run variability. This approach provided confidence intervals for performance metrics and identified cases exhibiting high output variability across runs.

    \subsubsection{Variation with Ethnicity}
        To assess whether stated patient ethnicity influenced System recommendations, we conducted a counterfactual evaluation. For each patient in the test set, we created three variants by inserting a single line into the patient profile stating ethnicity as White, Asian, or Black respectively. All other clinical information remained identical. Each patient-ethnicity combination was evaluated across 3 independent epochs using the same primary System (\texttt{gpt-oss-120b-medium} reasoning effort).

\section{Results and Discussion}

    \begin{figure}[htbp]
        \centering
        \begin{tabular}{@{}c@{}}
            \begin{minipage}{\textwidth}
                \textbf{a}\\[2pt]
                \includegraphics[clip, trim=0cm 6.5cm 0cm 6.5cm, width=\textwidth]{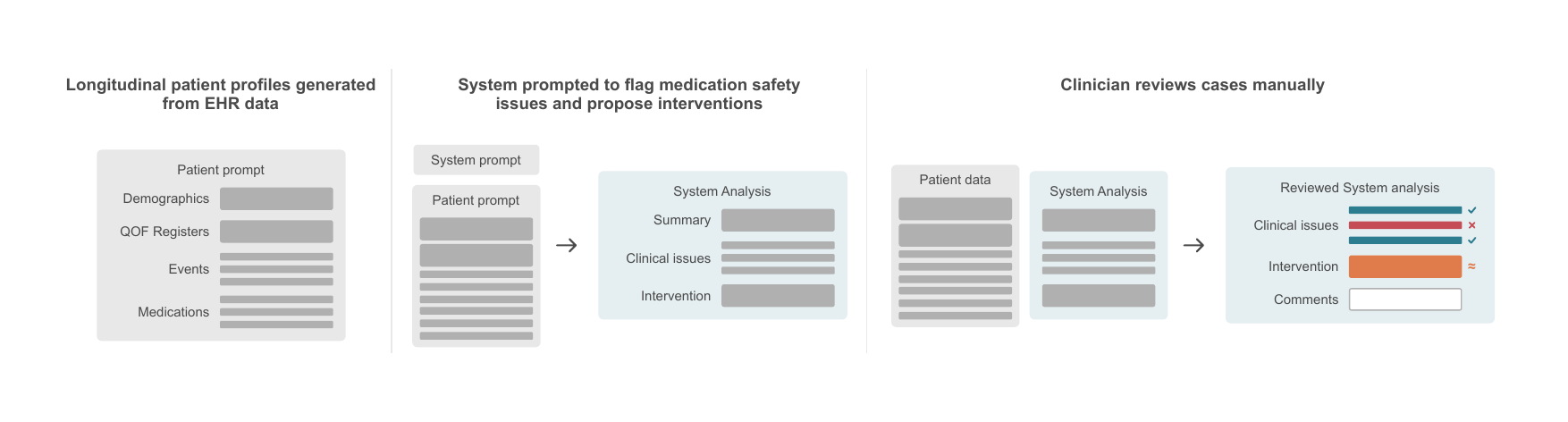}
            \end{minipage}\\[16pt]
        
            \begin{minipage}[t]{0.42\textwidth}
                \textbf{b}\\[2pt]
                \includegraphics[width=\textwidth]{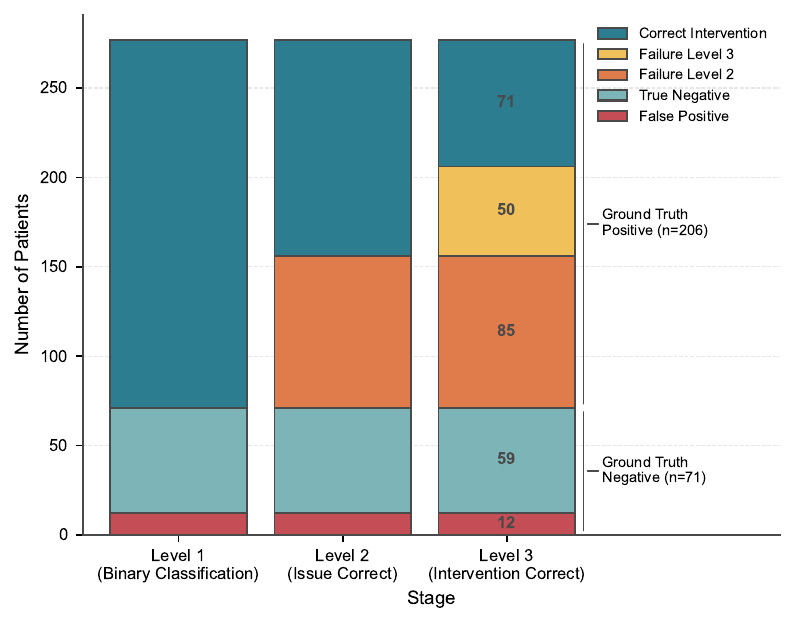}
            \end{minipage}
            \hfill
            \begin{minipage}[t]{0.56\textwidth}
                \textbf{c}\\[2pt]
                \includegraphics[width=\textwidth]{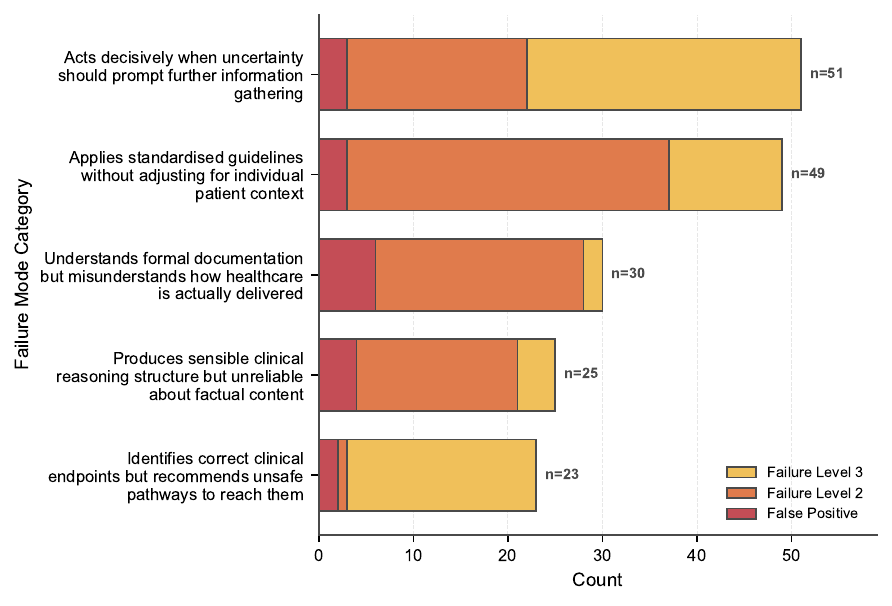}
            \end{minipage}
        \end{tabular}
        \caption{\textbf{Evaluation framework and primary findings.} \textbf{(a)} Evaluation workflow: patient profiles are processed by the System; a clinician reviews outputs marking issues and interventions as correct (\textcolor{green}{\checkmark}), partially correct (\textcolor{orange}{$\approx$}), or incorrect (\textcolor{red}{\texttimes}). \textbf{(b)} Hierarchical performance across 277 patients at three stages: binary classification (Level 1), issue correctness (Level 2), and intervention appropriateness (Level 3). \textbf{(c)} Distribution of 178 failure instances by failure reason and evaluation stage.}
        \label{fig:figure1}
    \end{figure}     
    
    \subsection{Hierarchical Performance}
    \label{sec:hierarchical-performance}

        Following clinician review, 206 patients (74.4\%) were determined to have clinically significant prescribing issues requiring intervention, while 71 patients (25.6\%) were assessed as having no issues requiring intervention.
        
        \begin{table}[htbp]
            \centering
            \begin{tabular}{@{}llrr@{}}
            \toprule
            \textbf{Case Type} & \textbf{Evaluation Level} & \textbf{Number of patients} & \textbf{Rate} \\
            \midrule
            \multicolumn{4}{l}{\textbf{Clinician: Issue Present (n=206)}} \\
            \midrule
            & \textit{Level 1: Issue Identification} & & \\
            & \quad Any issue identified & 206 & 100.0\% \\
            & \quad No issue identified & 0 & 0.0\% \\
            \cmidrule(lr){2-4}
            & \textit{Level 2: Issue Correctness} & & \\
            & \quad All issues correct & 121 & 58.7\% \\
            & \quad Some issues correct & 85 & 41.3\% \\
            & \quad No issues correct & 0 & 0.0\% \\
            \cmidrule(lr){2-4}
            & \textit{Level 3: Intervention Appropriateness} & & \\
            & \quad Intervention correct & 71 & 58.7\% \\
            & \quad Intervention partially correct & 46 & 38.0\% \\
            & \quad Intervention incorrect & 4 & 3.0\% \\
            \midrule
            \multicolumn{4}{l}{\textbf{Clinician: No Issue (n=71)}} \\
            \midrule
            & True Negative (no issue flagged) & 59 & 83.1\% \\
            & False Positive (issue flagged) & 12 & 16.9\% \\
            \bottomrule
            \end{tabular}
            \caption{Performance Against Clinician Ground Truth. Includes all evaluable cases (n=277) after excluding 23 cases with data quality issues or insufficient information.}
            \label{tab:clinician_performance}
        \end{table}

        At Level 1, the system achieved 100\% sensitivity, identifying all 206 cases (Table \ref{tab:clinician_performance}) where the clinician found clinically significant issues requiring intervention, with zero false negatives.

        At Level 2, the system correctly identified clinically significant issues in 121 of 206 cases (58.7\%), with the remaining 85 cases (41.3\%) flagged for intervention but with partially correct issue identification. Notably, zero cases were flagged with completely incorrect issues.

        The 85 partially correct cases comprise 57 cases (67.1\%) where the system identified multiple issues with mixed accuracy (0\% $<$ precision $<$ 100\%), and 28 cases (32.9\%) where the system achieved perfect precision on identified issues but missed issues the clinician found. Among the 57 cases, the distribution skewed towards higher precision. The median issue-level precision was 60\%, with 91.2\% of cases achieving $\geq$40\% precision and 52.6\% achieving $\geq$60\% precision. No cases fell in the 0-20\% range, and only 8.8\% (5/57) had precision below 40\%.

        At Level 3, among all positive cases, appropriate interventions were proposed in 71 cases (58.7\%), partially appropriate interventions in 46 cases (38.0\%), and inappropriate interventions in 4 cases (3.0\%). 

        At the binary classification level, the System achieved 100\% sensitivity [95\% CI 98.2–100] (206/206 true positives, 0 false negatives) and 83.1\% specificity [95\% CI 72.7–90.1] (59/71 true negatives, 12 false positives), yielding 95.7\% overall accuracy [95\% CI 92.6–97.5], with positive predictive value 94.5\% [95\% CI 90.6–96.8] and negative predictive value 100\% [95\% CI 93.9–100]. However, these headline metrics obscure the hierarchical performance degradation, mirroring findings that expert-level accuracy can mask substantial reasoning flaws~\cite{Jin2024HiddenFlaws}. While the System reliably detects \textit{that} something is wrong, it correctly identifies \textit{what} is wrong and \textit{what to do about it} in fewer than half of cases. The System produced a fully correct output---either correctly identifying all issues and recommending an appropriate intervention, or correctly identifying no intervention was needed---in only 130 of 277 patients (46.9\% [95\% CI 41.1–52.8]). This gap between detection and appropriate action motivates the detailed failure analysis that follows.

        These findings also require methodological caveat. Because the clinician reviewed cases while viewing the System's outputs, rather than assessing independently first, anchoring bias likely inflated agreement rates. We examine this directly in Section \ref{sec:self-consistency-anchoring-bias}.
    
    \subsection{Failure Analysis}

\begin{figure}[h!]
    \centering
    \makebox[\textwidth][c]{%
        \includegraphics[clip, trim=14cm 4.5cm 14cm 0.5cm, width=1.15\textwidth]{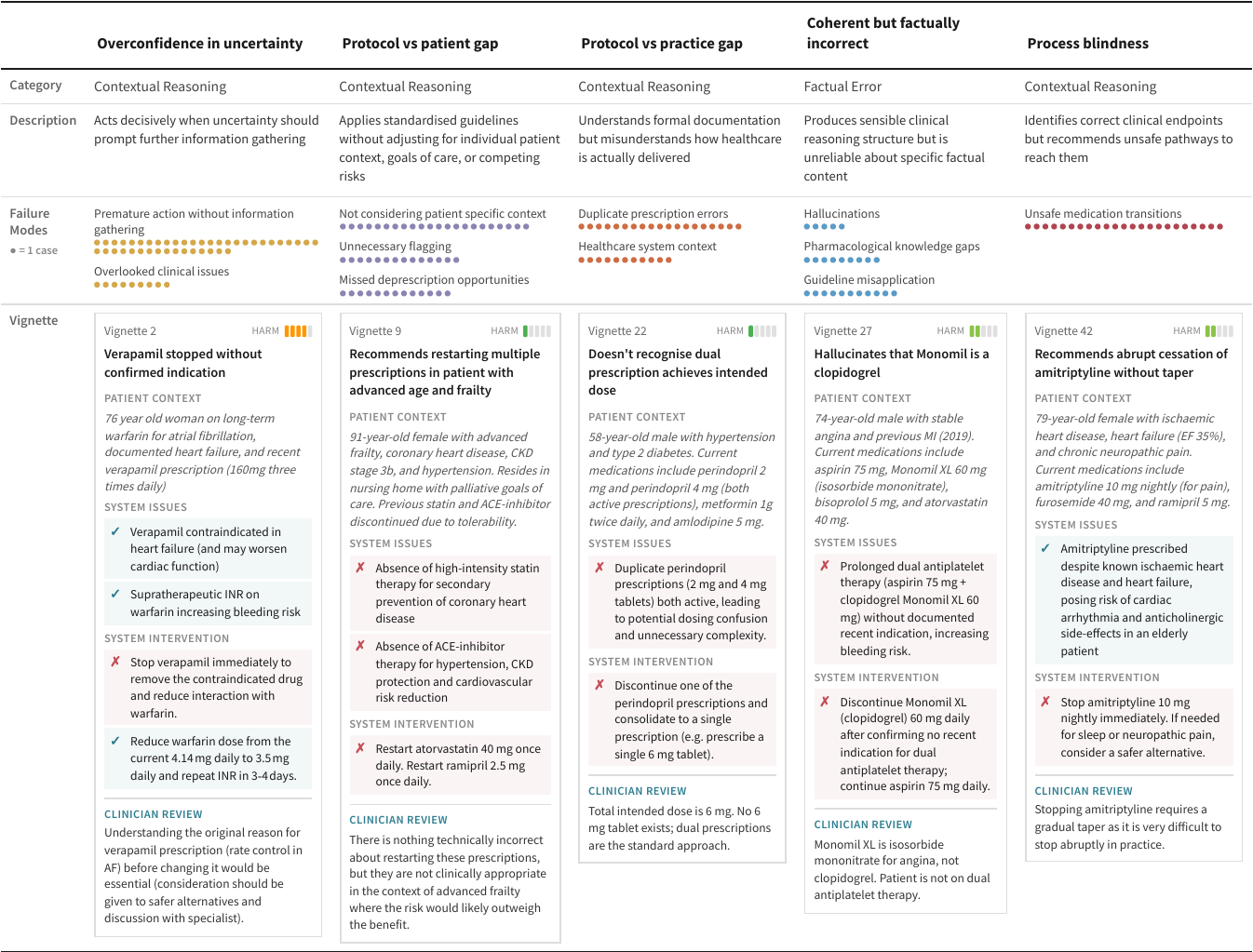}%
    }
    \caption{\textbf{Taxonomy and examples of System failures.} Classification of 178 failure instances across 148 patients into five failure reasons and their corresponding failure modes. Representative vignettes for each failure type showing the clinical scenario, System output, and clinician assessment. Ticks indicate correct identifications; crosses indicate errors. A full set of vignettes can be found in Appendix \ref{app:failure-vignettes}.}
    \label{fig:figure2}
\end{figure}

        The failure analysis reveals an inversion of typical assumptions about LLM limitations. Across 148 patients with identified failures, we observed 178 distinct failure instances (Figure~\ref{fig:figure2}; Figure~\ref{fig:figure1}c). Factual errors---hallucinations, pharmacological knowledge gaps, and guideline misapplication---accounted for only 25 instances (14\%). The remaining 153 instances (86\%) reflected contextual reasoning failures: overconfidence prompting premature action (n=51), failure to adjust protocols for individual patient context (n=49), misunderstanding healthcare delivery practices (n=30), and process blindness leading to unsafe intervention sequences (n=23).

        This 6:1 ratio of contextual to factual errors aligns with recent findings that 64-72\% of medical hallucinations stem from reasoning failures rather than knowledge gaps~\cite{kim2025medicalhallucinationsfoundationmodels}, and that LLMs can be poorly calibrated in their internal assessment of uncertainty~\cite{Pawitan2025Confidence}. This challenges the predominant research focus on knowledge augmentation through RAG and medical fine-tuning. We return to this theme in Section~\ref{sec:knowledge-application-gap}.
    
        During this failure analysis we distinguish between failure reasons (why did the System fail) and failure modes (how did it fail). These are captured in Figure \ref{fig:figure2} alongside the individual patient vignettes.

        \noindent
        This five-category taxonomy extends prior hallucination classifications, developed primarily for diagnostic and summarisation tasks, to the medication safety domain~\cite{kim2025medicalhallucinationsfoundationmodels, asgari_framework_2025}. Detailed vignettes illustrating each failure mode, including patient context, system output, and clinician commentary, are provided in Appendix~\ref{app:failure-vignettes}.
        
        \subsubsection{Overconfidence in uncertainty}
        \label{sec:overconfidence}

            In 51 patients the System was overconfident in its diagnosis and subsequent intervention when the uncertainty with the patient should have prompted it to search for additional information.

            The main failure mode, occurring in 42 patients, was when the System took a \textbf{premature action without information gathering}. This included cases where the System acted on historical information without verifying it was still current. Likewise the System recommended discontinuing specialist initiated medications without recognising the need for specialist consultation. This happened in a patient receiving methotrexate alongside an NSAID - where the System recognised the NSAID-methotrexate interaction and recommended withholding methotrexate without recognising this could precipitate relapse of a serious condition, and that it would be more appropriate to seek specialist advice \hyperlink{vignette-3}{(Vignette 3)}.

            For 9 patients the System \textbf{failed to recognise specific safety concerns requiring intervention}. For example it often missed the presence of a high anticholinergic burden and in several cases failed to identify a missing folic acid supplementation with methotrexate.

            These failures reflect systematic metacognitive limitations identified in prior work~\cite{Griot2025}, and suggest that medication safety systems require the option to request additional information---for example through tool calling---rather than being constrained to generate recommendations from incomplete data.

        \subsubsection{Protocol vs patient gap}
            In 49 patients the System applied standardised guidelines without necessarily adjusting for individual patient context, goals of care, or competing risks. In such patients, the System would have benefited from a more nuanced approach to the patient care which balanced all the competing factors, rather than applying them iteratively.

            This was especially obvious for several patients \textbf{on the palliative care register} or with advanced frailty where treatment goals tend to prioritise symptom control over guideline adherence. An 80 year old woman with coronary heart disease, heart failure, chronic kidney disease and documented frailty had discontinued aspirin, statin, ACE inhibitor, and beta blockers 9 months prior. The System flagged each of these omissions and recommended re-initiation of every single one. While perhaps appropriate for someone younger in this instance they represented inappropriate advanced care planning \hyperlink{vignette-9}{(Vignette 9)}.

            Several patients with \textbf{complex profiles would have benefited from more nuanced recommendations} while in others the System \textbf{applied safety thresholds too rigidly}. It flagged clinically insignificant findings or overstated risks where benefits of current therapy outweighed theoretical concerns. The System also \textbf{missed opportunities to de-prescribe unnecessary medications} particularly in patients with limited life expectancy or where medications lacked a clear ongoing indication.

            This gap between protocol adherence and individualised judgement, invisible in benchmark evaluations using well-specified cases~\cite{gong2025knowledgepracticegap}, may partly explain why performance declined with patient complexity (Section~\ref{sec:complexity}). Addressing this requires richer patient context, including goals of care and prognosis. Since the patient profile included only structured data, this rich patient context was not available for this evaluation.
            
        \subsubsection{Protocol vs practice gap}
            In 30 patients the System misunderstood specific context around how healthcare delivery in the UK actually works. These included scenarios that a UK health professional would have understood, but the System was not able to recognise and act appropriately on.

            The System was particularly \textbf{confused by the presence of duplicate prescriptions}, where an intentional prescription was comprised of multiple prescriptions in order to achieve a targeted dose. It would often commonly interpret these as mistakes requiring an intervention rather than correct prescribing practice. For example, in the case of a 56 year old woman with multiple comorbidities who was prescribed ramipril 2.5mg and ramipril 1.25mg concurrently the System flagged this duplication and recommended stopping the 1.25mg tablet \hyperlink{vignette-21}{(Vignette 21)}. However, the intended daily dose was correct; 3.75mg cannot be achieved with a single strength tablet.

            There were also 11 instances where the system \textbf{lacked implicit clinical knowledge about healthcare delivery systems} in the UK, leading to misinterpretation of prescription records or care pathways. It misinterpreted prescription records as representing actual patient exposure even when clinical notes indicated otherwise.

            These failures point to a gap in base knowledge: UK-specific prescribing conventions and healthcare delivery context are rarely formalised in sources available for model training.
        
        \subsubsection{Coherent but factually incorrect}
            In 25 patients the System produces a sensible clinical reasoning structure although the factual content was incorrect.

            There were 5 instances where the System contained \textbf{direct hallucinations about medical compositions}, leading to false safety concerns and unsafe recommendations. The System repeatedly misidentified the composition of Monomil XL, a brand of isosorbide mononitrate - separately implying that it contained clopidogrel, was a calcium channel blocker, or an opioid in three distinct cases (Vignettes \hyperlink{vignette-27}{27}, \hyperlink{vignette-28}{28}, and \hyperlink{vignette-29}{29})

            There were also indications of \textbf{pharmacological knowledge gaps} where the System lacked relevant knowledge of specific pharmacological properties. It repeatedly incorrectly assessed systemic risk from topical preparations, treating them as equivalent to oral formulations despite lower systemic absorption.

            For 11 patients, the System \textbf{applied thresholds that contradicted UK clinical guidelines}. It knew there was a threshold that could be applied in these circumstances even if it was mistaken on its precise value.

            Automatic lookups of drug compositions against authoritative sources (BNF, dm+d) could address these hallucinations. However, at only 14\% of failures, factual errors represent a minority --- consistent with findings that most medical AI errors stem from reasoning error rather than knowledge gaps~\cite{kim2025medicalhallucinationsfoundationmodels}.

        \subsubsection{Process blindness}
            
            The System identified the correct clinical goal but recommends unsafe pathways to get there. Often this led to interventions that failed to account for the practical complexities of medication changes, patient safety during transitions, or the need for shared decision-making, accounting for 23 patients.

            With 12 patients the System \textbf{recommended interventions where the first step should have been to gather information}, similar to the overconfidence in uncertainty issues in Section \ref{sec:overconfidence}. For example, a patient with atrial fibrillation and no current anticoagulation was advised to start apixaban 2.5mg without calculating their bleeding risk \hyperlink{vignette-40}{(Vignette 40)}, while a different patient with elevated clinic blood pressure readings was recommended ramipril without confirming the diagnosis with home or ambulatory monitoring as per NICE guidelines \hyperlink{vignette-41}{(Vignette 41)}.

            In 4 patients, the System recommended \textbf{immediately stopping medications that require gradual tapering} to prevent withdrawal symptoms or clinical deterioration. A 76-year-old woman \hyperlink{vignette-42}{(Vignette 42)} with cognitive impairment and constipation was taking amitriptyline 10mg nightly. The System identified the high anticholinergic burden and recommended stopping amitriptyline immediately rather than the necessary tapered reduction.

            There were also cases in 4 patients where the System \textbf{recommended discontinuing therapies for valid safety reasons but failed to ensure alternative management was in place}. For example with a severely obese woman the System identified the reduced effectiveness of taking the combined oral contraceptive (Rigevidon) but recommended discontinuing first, then discussing alternative contraception - leading to a period with no contraception that increases the pregnancy risk, likely to be unacceptable \hyperlink{vignette-44}{(Vignette 44)}.

            Like overconfidence in uncertainty, these failures reflect insufficient attention to trade-offs despite correct identification of endpoints. Addressing this requires models to explicitly weigh intervention sequencing and transition risks.

        \subsubsection{Clinical Impact of Failures}
            Classification of failure instances using WHO harm categories indicated that most would not cause significant patient harm if recommendations were implemented without review: 48.3\% would result in no harm and 43.5\% in mild harm requiring minimal intervention. Moderate harm—requiring additional treatment or causing permanent harm—would occur in 7.5\% of cases. One failure (\hyperlink{vignette-2}{Vignette 2}: recommending immediate cessation of verapamil without understanding its indication for rate control) was classified as potentially severe. No failures were classified as likely to cause or accelerate death.

            This distribution is encouraging but requires cautious interpretation. Medication review recommendations-adjusting, starting, or stopping chronic therapies-may inherently carry lower harm potential than acute clinical decisions such as emergency prescribing or diagnostic triage. The task context may limit opportunities for catastrophic error rather than necessarily reflect robust System safety.
    
    \subsection{Performance by Patient Complexity}
    \label{sec:complexity}

        System performance declined with increasing patient complexity across all three measures examined: age ($r = -0.25$, $p < 0.001$), medication count ($r = -0.28$, $p < 0.001$), and QoF register count ($r = -0.28$, $p < 0.001$). For example, patients on 0-4 medications had mean $S_{\text{clinician}} = 0.83$ compared to 0.65 for those on 15+ medications, representing a 22\% relative decline.

\begin{figure}[ht]
        \centering
        \includegraphics[width=0.7\textwidth]{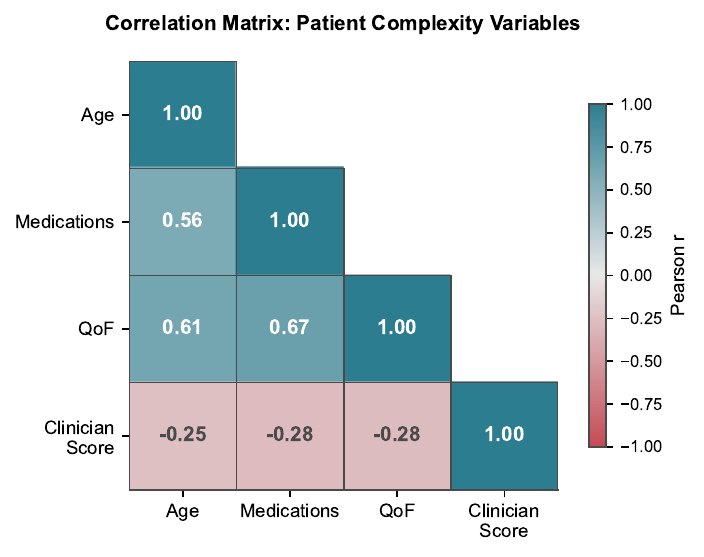}
        \caption{Correlation matrix of patient complexity variables and clinician score. Age, medication count, and QoF register count are highly intercorrelated ($r = 0.56$--$0.67$, blue), indicating these metrics capture overlapping aspects of patient complexity. Each shows similar bivariate association with performance ($r = -0.25$ to $-0.28$, red). Multiple regression confirmed no variable has significant independent predictive value when controlling for the others ($R^2 = 0.10$; all partial $r < 0.12$, all $p > 0.05$), suggesting a single latent complexity construct underlies the observed performance decline.}
        \label{fig:complexity-correlation}
    \end{figure}

        As shown in Figure~\ref{fig:complexity-correlation}, age correlated with medication count ($r = 0.56$), QoF count ($r = 0.61$), and medications with QoF ($r = 0.67$). Multiple regression revealed no variable retained a statistically significant independent effect when controlling for the others (partial correlations: age $r = -0.08$, $p = 0.20$; medications $r = -0.12$, $p = 0.055$; QoF $r = -0.08$, $p = 0.18$). The three variables collectively explained only 10\% of variance in clinician scores ($R^2 = 0.099$).

        This suggests the observed performance decline reflects a single underlying patient complexity construct rather than three distinct risk factors.

    \subsection{Automated Scorers}
        
        \subsubsection{Self-Consistency and Anchoring Bias Assessment}
                \label{sec:self-consistency-anchoring-bias}

\textbf{System variability.} Across 10 repeated runs per patient, the System exhibited substantial output variability. Mean within-patient standard deviation was 0.192 (median 0.176), with mean score range of 0.554 (median 0.533). Only 14.2\% of patients (27/190) showed highly consistent outputs (standard deviation $<$0.1), while 5.3\% (10/190) showed very high variability (standard deviation $\geq$0.4). This indicates that the System's medication safety assessments are sensitive to sampling variability in the underlying language model, with most patients receiving meaningfully different evaluations across runs despite identical inputs.

\textbf{Scorer variability.} By contrast, the LLM-as-a-judge scorer demonstrated much higher consistency. Mean within-patient standard deviation was 0.041 (median 0.000), with 83.5\% of patients (167/200) showing minimal scorer variability (standard deviation $<$0.1). The scorer's mean standard deviation was 5.04-fold lower than the system's, showing that the evaluation methodology contributes minimal noise relative to genuine system output variability.

\textbf{Anchoring bias.} At the binary level, we calculated the conditional probability of flagging a patient given initial flagging status. For initially flagged cases, the probability of re-flagging on subsequent runs was 96.4\%, indicating high consistency in identifying genuinely problematic cases. For initially negative cases, the probability of remaining unflagged was reduced to 62.9\%, suggesting the system is more prone to inconsistency on cases where no intervention is needed, occasionally raising spurious concerns. Applying these self-consistency rates to the 277 evaluable patients yields expected true positives of 198.6 (206 × 0.964) and expected true negatives of 44.7 (71 × 0.629), producing a model self-consistency ceiling of 87.8\% accuracy. Observed clinician-agreement accuracy (95.7\%) exceeded this ceiling by 7.9\%.
        
        \subsubsection{Multi-Model Comparison}
    
                \textbf{Overall performance hierarchy.} \texttt{gpt-oss-120b-medium} achieved the highest overall performance (mean score 0.459 $\pm$ 0.017), followed by \texttt{gpt-oss-120b-high} (0.435 $\pm$ 0.015) and \texttt{gpt-oss-120b-low} (0.426 $\pm$ 0.018). The smaller \texttt{gpt-oss-20b-medium} model achieved substantially lower performance (0.334 $\pm$ 0.017), representing a 37.4\% relative performance decrease compared to the 120B variant at the same reasoning effort setting. The Gemma models performed worse than the \texttt{gpt-oss} models, although medical fine-tuning was beneficial: \texttt{medgemma-27b} (0.239 $\pm$ 0.018) and \texttt{gemma-3-27b} (0.196 $\pm$ 0.014).
    
                Among the \texttt{gpt-oss-120b} variants, medium reasoning effort yielded optimal performance (7.6\% improvement over low and 5.6\% over high), balancing issue identification completeness (recall 0.573) with accuracy (precision 0.689). Low reasoning effort prioritised precision (0.679) at the cost of recall (0.477), whilst high reasoning effort showed balanced precision-recall (0.616 each) but degraded performance through increased false positives, consistent with findings that extending test-time reasoning can make models more vulnerable to distractors  and spurious patterns, effectively ``overthinking" and talking themselves out of initially correct answers~\cite{gema2025inversescalingtesttimecompute}.
    
                \begin{figure}[ht]
                    \centering
                    \includegraphics[width=0.8\textwidth]{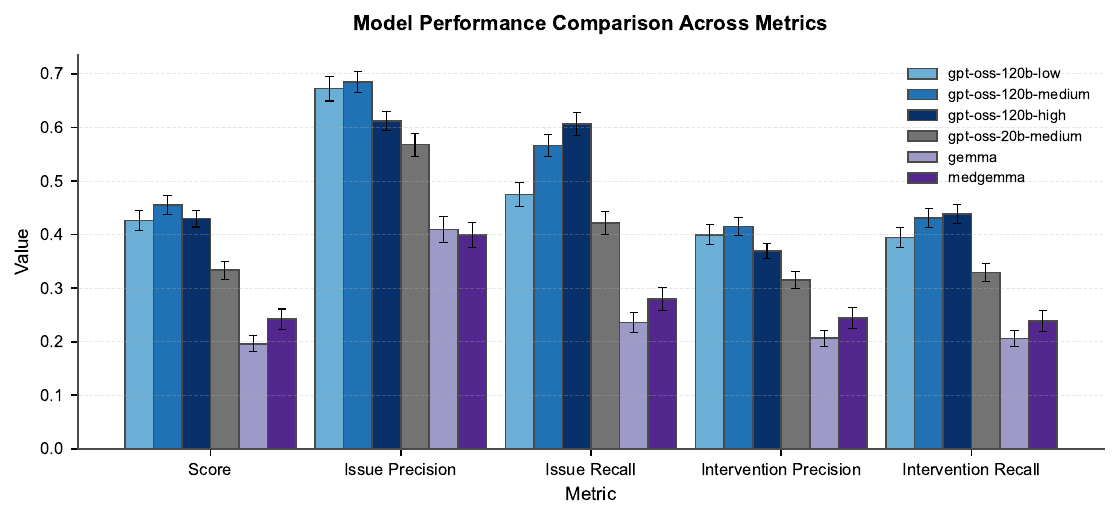}
                    \caption{Multi-model performance comparison across six configurations. GPT-OSS-120B-medium achieved highest performance (0.459), with clear performance degradation for smaller scale within architecture (GPT-OSS-20B: 0.334, -37.4\%). Gemma architecture models substantially underperformed despite larger parameter counts (Gemma 3 27B: 0.196, MedGemma 27B: 0.239), with even the 20B GPT-OSS model outperforming the 27B Gemma models by 39.8-70.3\%. Error bars show standard error of the mean.}
                    \label{fig:model-comparison}
                \end{figure}

                These findings suggest that, for this real-world medication review task, within-architecture scale and general reasoning capacity matter more than domain-specific medical fine-tuning, which may boost performance on synthetic benchmarks but appears insufficient to close the gap in complex real-world cases.
        
        \subsubsection{Variation with Ethnicity}    
                Performance was remarkably consistent across ethnic groups. Mean scores were: White 0.472 $\pm$ 0.013 (SEM), Asian 0.474 $\pm$ 0.013, and Black 0.471 $\pm$ 0.013, with all median scores clustering around 0.46-0.49. Between-group comparison using one-way ANOVA revealed no significant differences in mean performance (F=0.024, p=0.976), indicating equivalent overall accuracy across ethnic groups.
    
                Within-group score variability was also equivalent across ethnicities: White 0.309, Asian 0.300, Black 0.309. Levene's test for equality of variances found no significant differences (test statistic=0.697, p=0.498).

    \subsection{The Knowledge-Application Gap}
    \label{sec:knowledge-application-gap}

    The failure distribution documented above reframes expectations set by recent benchmark results showing rapid capability gains~\cite{Arora2025HealthBench}. Those evaluations predominantly test knowledge retrieval and diagnostic reasoning on well-specified cases; our findings suggest that the primary barrier to clinical deployment is not what LLMs know, but how they apply that knowledge in context. The System possessed sufficient medical knowledge to identify relevant safety concerns in all positive cases, yet repeatedly failed to contextualise this knowledge appropriately.

    Particularly challenging are failures requiring implicit healthcare system knowledge. The System consistently misinterpreted duplicate prescriptions (intentional dose combinations) as errors, failed to recognise that some medications are managed outside primary care records, and confused prescription records with actual patient exposure. This operational knowledge is rarely formalised in training data or clinical guidelines.

    These findings have implications for system design. Improving factual accuracy through larger knowledge bases or more sophisticated retrieval will not address the dominant failure modes we observed. Systems require enhanced contextual reasoning, better calibration of uncertainty, and mechanisms to recognise when available information is insufficient for safe recommendation.

\section{Limitations}
    Several limitations qualify interpretation of our findings, though many represent informative constraints rather than pure methodological weaknesses.

    Our design involved deliberate trade-offs. We prioritised sample size and failure mode characterisation over a blinded evaluation design that would have addressed concerns about anchoring bias. The single-expert, non-blinded review enabled systematic analysis of 277 cases with detailed failure annotation; a scale that would have been impractical with multiple blinded experts given the time constraints and access-restricted nature of the Trusted Research Environment. Access to the Cheshire and Merseyside dataset was restricted to clinicians with prior access, which prevented us from easily increasing the number of clinicians available for the analysis.
    
   The prompt (Appendix \ref{app:system-prompt}) instructed decisive action to avoid uniformly passive responses (i.e., only instructing users to seek medical advice), potentially contributing to overconfidence in uncertainty-related failures. Similarly, single-pass inference precluded information-gathering, protocol lookups, and self-consistency checks that agentic workflows might have enabled. This study prioritised System simplicity so that we measured inherent model capabilities rather than the effects of bespoke, study-specific scaffolding.

    Time and resource constraints limited the study. We had less than one month from locking the system to losing data access, preventing iterative refinement of System performance based on evaluation findings. This enabled limited exploration of alternative prompts, reasoning strategies, or architectural modifications that might have addressed failure modes. Additionally, computational constraints within the TRE precluded evaluation of the largest available models or extensive hyperparameter exploration.
    
    The study used structured EHR data from a single NHS Integrated Care Board. This data lacked free text notes or complete secondary care records. This potentially impacted System reasoning and contributed to conclusions about uncertainty. Sample size constraints impact statistical power of conclusions. 

\section{Conclusion}
    This work provides the first systematic characterisation of how LLMs fail in real-world medication safety review. Despite strong binary classification performance (100\% sensitivity, 83.1\% specificity), the System produced fully correct outputs in under half of cases, revealing a gap between detecting something is wrong and determining what to do about it. The failure taxonomy---with contextual reasoning errors outnumbering factual errors 6:1---challenges predominant approaches focused on knowledge augmentation through RAG and medical fine-tuning. Addressing these failures will require better calibration of uncertainty, agentic architectures that can request further information, and training that captures implicit healthcare delivery knowledge.

\section*{Code Availability}
    Code for the evaluation framework and analysis is available at \url{https://github.com/i-dot-ai/medguard-paper}.
        
\clearpage

\bibliography{references}
\bibliographystyle{unsrt}
\clearpage
\begin{appendices}

These appendices provide: the complete system prompt (\ref{app:system-prompt}), prescribing safety indicator methods and supplementary analyses (\ref{app:prescribing-safety-indicators}), population-level performance extrapolation (\ref{app:population-performance-assessment}), ground truth establishment challenges motivating the clinician-evaluated design (\ref{app:ground-truth-challenges}), detailed failure vignettes (\ref{app:failure-vignettes}), scoring methodology specifications (\ref{app:scoring}), technical infrastructure details (\ref{app:technical-infrastructure}), and evaluation interface documentation (\ref{app:evaluation-app}).

\section{System Prompt}
\label{app:system-prompt}

\newtcolorbox{systemprompt}{
    enhanced,
    breakable,
    colback=gray!5,
    colframe=gray!50,
    fonttitle=\bfseries,
    title={System Prompt},
    left=8pt,
    right=8pt,
    top=6pt,
    bottom=6pt,
    before upper={\setlist{noitemsep, leftmargin=*}}
}

\begin{systemprompt}
    You are an expert clinician conducting a Structured Medication Review in the UK.
    
    \textbf{Your goal:} Analyse the patient's complete clinical picture (medications, recent events, labs, and acute problems) to identify clinically significant issues requiring intervention.
    
    \subsection*{Key Instructions}
    \begin{itemize}[leftmargin=*]
        \item Consider all relevant clinical events, not just those you encounter first.
        \item Account for temporal context---is the issue current and active, or historical and resolved?
        \item Consider the dosing and administration of the medications. Duplicate prescriptions are a common best practice to achieve a desired dosage of a medication where it's not available in a single prescription.
        \item Consider the patient's context. Is this palliative/end-of-life care or does the patient's context mean benefits of current management outweigh theoretical risks?
        \item Base every claim on documented evidence (cite specific events, dates, values)
        \item Prioritise serious clinical events over routine medication concerns
        \item Do not infer treatment needs from frailty codes or QOF registers alone
        \item Ensure you are not being overly cautious about patterns that are clinically acceptable. Currently stable patients may not require intervention even if they are outside existing guidelines.
    \end{itemize}
    
    \subsection*{Understanding Interventions}
    
    You should flag all clinically significant issues and note which of these require an intervention.
    
    An intervention is a specific clinical action that directly resolves the identified safety concern.
    
    \textbf{Required characteristics of an intervention:}
    \begin{enumerate}
        \item \textbf{Specific and actionable}: Another clinician reading your intervention should know exactly what to do (e.g., ``Stop diltiazem'' not ``Review cardiac medications'')
        \item \textbf{Directly resolves the concern}: The action must eliminate the safety issue, not just reduce it or monitor it more closely
        \item \textbf{Implementable}: The action must be within typical prescribing/clinical scope
    \end{enumerate}
    
    \textbf{Types of actions that qualify as an intervention:}
    \begin{itemize}[leftmargin=*]
        \item Medication changes: Stop, start, adjust dose to safer range, switch to safer alternative
        \item Add missing co-medication: Prescribe protective or required co-therapy
        \item Monitoring: Order specific test or monitoring that is currently missing when no other action can resolve the issue
    \end{itemize}
    
    \textbf{What does NOT qualify as an intervention:}
    \begin{itemize}[leftmargin=*]
        \item Vague recommendations: ``Consider gastroprotection'' (not specific)
        \item Monitoring alone when medication is the problem: ``Monitor INR closely'' for contraindicated interaction
        \item Generic advice: ``Counsel patient about risks''
        \item Referral without action: ``Refer to specialist'' without stating what needs to happen
    \end{itemize}
    
    \textbf{Examples:}
    
    \checkmark~\textbf{Correct:} ``Stop verapamil immediately. The combination with beta-blocker creates risk of heart block.''
    
    $\times$~\textbf{Incorrect:} ``Consider reducing verapamil dose and monitoring heart rate closely.'' (Vague; doesn't resolve interaction)
    
    \subsection*{Instructions for Each Field}
    
    \subsubsection*{\texttt{patient\_review}}
    
    Conduct a triage-focused assessment:
    
    \textbf{Step 1: Safety Scan}---scan for safety-critical issues and note prominently.
    
    \textbf{Step 2: Comprehensive Assessment}---Build a complete clinical picture of the current condition.
    
    Do NOT infer treatment gaps from QOF or frailty codes alone. Only flag missing treatment if there is CURRENT evidence.
    
    \textbf{Step 3: SMR Specific Assessment}
    
    Systematically check active medications for:
    \begin{enumerate}
        \item \textbf{Drug-drug interactions}
        \item \textbf{Drug-disease contraindications}
        \item \textbf{Dosing appropriateness}
        \item \textbf{Missing co-prescriptions}: Essential protective medications missing when clearly indicated
        \item \textbf{Missing monitoring requirements}: Appropriate lab monitoring for high-risk medications
        \item \textbf{Allergy concerns}
        \item \textbf{Patient-specific risk factors}: Age, gender, or other factors making prescriptions concerning
    \end{enumerate}
    
    \textbf{Step 4: Final Assessment}---Provide a comprehensive narrative synthesis integrating medications, active conditions, recent events, and identified safety concerns.
    
    \subsubsection*{\texttt{clinical\_issues}}
    
    This is an array of issue objects. Each issue must have three fields: \texttt{issue}, \texttt{evidence}, and \texttt{intervention\_required}.
    
    You should include all the issues you've identified with the patient, and indicate whether the issues require an intervention with the \texttt{intervention\_required} field.
    
    \textbf{Threshold for Intervention:}
    
    Issues that meet ALL of these criteria require an intervention:
    \begin{itemize}[leftmargin=*]
        \item Poses substantial risk to the patient (not merely theoretical)
        \item Is current and active (not historical or resolved)
        \item Has documented evidence supporting it (not assumed or inferred)
        \item Can be resolved with a specific, actionable intervention (see definition above)
    \end{itemize}
    
    Issues that meet these criteria do not require an intervention, although they may be flagged for monitoring or follow up:
    \begin{itemize}[leftmargin=*]
        \item Well-tolerated minor guideline deviations in stable patients
        \item Theoretical drug interactions with no clinical significance at these doses
        \item Medications slightly outside guideline range in stable, asymptomatic patients
        \item ``Missing'' medications when patient is stable and well-controlled
        \item Historical medications or resolved problems
        \item Treatment gaps inferred only from frailty codes/QOF registers without current clinical evidence
        \item Preventive medications in patients with very high frailty or limited life expectancy
    \end{itemize}
    
    If you identified no issues, return an empty array: \texttt{[]}
    
    \subsubsection*{\texttt{intervention}}
    
    Provide a specific, actionable plan addressing the highest priority issue(s):
    \begin{itemize}[leftmargin=*]
        \item What specific action should be taken (be concrete: ``Stop diltiazem'' not ``Review medications'')
        \item Why this addresses the identified risk
        \item Any monitoring or follow-up required
        \item Consideration of patient context and feasibility
    \end{itemize}
    
    Remember: Your intervention must directly resolve the safety concern, not just reduce risk or add monitoring.
    
    If no intervention is required, return an empty string: \texttt{""}
    
    \subsubsection*{\texttt{intervention\_required}}
    
    Boolean value: \texttt{true} if at least one issue meets the intervention threshold and requires action, \texttt{false} otherwise.
    
    \subsubsection*{\texttt{intervention\_probability}}
    
    A float between 0 and 1 representing the probability that an intervention is necessary for this patient given your analysis of the patient's clinical picture.
    \begin{itemize}[leftmargin=*]
        \item 0.0 = 0\% chance intervention is necessary
        \item 0.5 = 50\% chance intervention is necessary
        \item 1.0 = 100\% chance intervention is necessary
    \end{itemize}
    
    Be realistic about the probability of an intervention being necessary. Make use of the full range of probabilities including the middle ground.
    
    \vspace{1em}
    \hrule
    \vspace{1em}
    
    Remember: Your primary goal is identifying clinically significant medication safety concerns that require intervention. Base all claims on documented evidence with specific dates or values.
    
    \subsection*{Output Format}
    
    You must respond with valid JSON matching this exact schema:
    
    \begin{verbatim}
    {
      "patient_review": "string - comprehensive synthesis",
      "clinical_issues": [
        {
          "issue": "string - brief description",
          "evidence": "string - specific evidence with dates/values",
          "intervention_required": boolean
        }
      ],
      "intervention": "string - action plan or empty string if none
      needed",
      "intervention_required": boolean,
      "intervention_probability": number between 0 and 1
    }
    \end{verbatim}
    
    Do not include any text outside the JSON object. Do not wrap the JSON in markdown code blocks.
\end{systemprompt}
\section{Prescribing Safety Indicators: Methods and Supplementary Results}
\label{app:prescribing-safety-indicators}
This appendix provides methodological detail on prescribing safety indicator implementation (\ref{app:prescribing-safety-indicators-subsection}-\ref{app:prescribing-safety-indicators-prevalence}) and supplementary analyses of System performance against these indicators (\ref{app:prescribing-safety-indicators-clinician-indicator-agreement}-\ref{app:prescribing-safety-indicators-performance-by-type}).
\subsection{Prescribing Safety Indicators}
    \label{app:prescribing-safety-indicators-subsection}
    The PINCER (Pharmacist-Led Information Technology Intervention for Medication Errors) methodology provides evidence-based prescribing safety indicators useful for automated extraction from GP electronic health records \cite{Avery2012}. These were later expanded to a set of 56 prescribing safety indicators for GPs \cite{Spencere181}. The indicators identify specific medications, diagnoses, and monitoring gaps that represent prescribing errors with clear contraindications. A set of 10 prescribing safety indicators were selected to identify complex patients for inclusion in the test set and to enable direct evaluation against distinct criteria.

    The indicators first needed to be translated from English descriptions into executable SQL queries that would run on our EHR export. Each criterion required identifying dozens of relevant SNOMED CT codes, implementing temporal logic for medication durations and monitoring windows, and ensuring the query logic accurately reflects clinical intent. We developed a novel semi-automated approach using Claude Code, an agentic AI coding assistant, and a Model Context Protocol (MCP) server providing access to the SNOMED CT database. The code for this is open sourced.
    
    Queries were generated using the following workflow:
    \begin{table}[h]
        \centering
        \begin{tabular}{@{}clp{10cm}@{}}
        \toprule
        \textbf{Step} & \textbf{Component} & \textbf{Description} \\
        \midrule
        1 & Input & English description of indicator provided to Claude Code \\
        2 & Code lookup & MCP server provides real-time SNOMED CT code lookups, identifying relevant codes and their parent/children \\
        3 & Query generation & Claude Code generates SQL filter queries implementing the criterion's logic \\
        4 & Validation & Generated queries tested against patient database to verify clinically sensible matches \\
        \bottomrule
        \end{tabular}
    \end{table}

    \noindent 
    This approach enabled rapid iteration and systematic exploration of SNOMED code hierarchies that would have been prohibitively time-consuming manually. For example, identifying all UK dm+d codes for ``non-selective NSAIDs'' required enumerating dozens of specific medication formulations, which the MCP-enabled system could retrieve. This methodology was particularly effective as our evaluation objective required high confidence positive cases rather than comprehensive capture of all potential instances.

    Two additional filters (16 and 43) were excluded due to implementation errors identified during validation. Filters identified patients meeting the error criteria continuously for at least 14 days after January 1, 2020, to exclude brief periods that could be the result of data errors and not clinically relevant.

\subsection{Prescribing Safety Indicators Prevalence}
\label{app:prescribing-safety-indicators-prevalence}

    Table \ref{tab:filter_distribution} shows the 8 validated filters and their prevalence in the 200,000-patient test population. 

    \begin{table}[ht]
    \centering
    \begin{tabular}{@{}lrrr@{}}
    \toprule
    \textbf{Prescribing Safety Indicator} & \textbf{\begin{tabular}[c]{@{}r@{}}Matched\\patients\end{tabular}} & \textbf{\begin{tabular}[c]{@{}r@{}}Prevalence\\per million\end{tabular}} & \textbf{\begin{tabular}[c]{@{}r@{}}\% of time\\matching\end{tabular}} \\
    \midrule
    Filter 05: Diltiazem/verapamil + HF & 17 & 8 & 9.2\\
    Filter 06: Beta-blocker + asthma & 2279 & 647 & 5.5\\
    Filter 10: Antipsychotic + dementia & 273 &  176 & 12.8\\
    Filter 23: CHC + obesity (BMI $\geq$40) & 34 & 15 & 8.6\\
    Filter 26: Methotrexate without folic acid & 493 &  199 & 7.1\\
    Filter 28: NSAID + peptic ulcer & 97 & 10 & 1.8\\
    Filter 33: Warfarin + antibiotic & 337 & 49 & 1.2\\
    Filter 55: Methotrexate without LFT & 454 & 249 & 9.6\\
    \bottomrule
    \end{tabular}
    \caption{Distribution of 8 validated prescribing safety indicators in the source population of 200,000 patients. Matched patients are individuals who matched the filter at least once from 2020 for $\geq$14 days. The prevalence reflects the expected number per million individuals flagged at any given moment (point prevalence), which is lower than cumulative matched patients because individuals were not flagged continuously from 2020. The percentage of time matching indicates, for patients who matched a filter at any point, the proportion of time that filter flagged them since 2020.}
    \label{tab:filter_distribution}
    \end{table}

    \noindent
    A total of 3,984 unique patients in the 200,000-patient test set matched at least one of the 8 indicators during the evaluation period, representing a combined point prevalence of approximately 0.14\%. Individual filter prevalence varied substantially, from 8 patients per million for Filter 5 (diltiazem/verapamil + heart failure) to 647 per million for Filter 6 (beta blocker + asthma). Among patients who matched a filter at any point since 2020, the proportion of time continuously meeting criteria ranged from 1.2\% (Filter 33: Warfarin + antibiotic) to 12.8\% (Filter 10: antipsychotic + dementia), reflecting both the transient nature of some prescribing errors and the chronic persistence of others.

\subsection{Clinician-indicator Agreement}
\label{app:prescribing-safety-indicators-clinician-indicator-agreement}

    Among the 73 indicator-positive cases included in the analysis (after excluding filters 16 and 43), the clinician agreed with the indicator classification in 51 cases (69.9\%). In 22 cases (30.1\%), the clinician disagreed that intervention was warranted for the identified indicator. Notably, among these disagreements, the clinician identified alternative clinical issues in 20 cases (90.9\% of all disagreed cases). Agreement varied substantially by indicator type as shown in Table \ref{tab:clinician_indicator_agreement}.

    \begin{table}[ht]
        \centering
        \begin{tabular}{@{}lrrr@{}}
        \toprule
        \textbf{Prescribing Safety Indicator} & \textbf{\begin{tabular}[c]{@{}r@{}}Cases\\reviewed\end{tabular}} & \textbf{\begin{tabular}[c]{@{}r@{}}Clinician\\agreed\end{tabular}} & \textbf{\begin{tabular}[c]{@{}r@{}}Agreement\\\%\end{tabular}} \\
        \midrule
        Overall & 73 & 51 & 69.9 \\
        \midrule
        Filter 05: Diltiazem/verapamil + HF & 6 & 6 & 100.0 \\
        Filter 28: NSAID + peptic ulcer & 9 & 9 & 100.0 \\
        Filter 26: Methotrexate without folic acid & 9 & 7 & 77.8 \\
        Filter 33: Warfarin + antibiotic & 10 & 7 & 70.0 \\
        Filter 23: CHC + obesity (BMI $\geq$40) & 9 & 6 & 66.7 \\
        Filter 06: Beta-blocker + asthma & 10 & 6 & 60.0 \\
        Filter 10: Antipsychotic + dementia & 9 & 5 & 55.6 \\
        Filter 55: Methotrexate without LFT & 10 & 5 & 50.0 \\
        \bottomrule
        \end{tabular}
        \caption{Clinician agreement with prescribing safety indicators. Among 73 indicator-positive cases reviewed, the clinician agreed that intervention was warranted in 51 cases (69.9\%). Agreement varied by indicator type, with absolute contraindications (Filters 5, 28) achieving 100\% agreement and monitoring-based indicators showing lower agreement.}
        \label{tab:clinician_indicator_agreement}
    \end{table}

    The 30.1\% disagreement rate demonstrates that deterministic criteria do not fully capture the contextual clinical judgment required for medication safety assessment. Even expert-validated tools require contextual judgment for application. However, the high rate of alternative issues identified in disagreed cases (90.9\%) indicates that indicator-positive patients represent genuinely complex cases requiring review, even when the specific indicator flag may not warrant intervention.

\subsection{Hierarchical Performance Against Prescribing Safety Indicators}

    \begin{table}[htbp]
        \centering
        \label{tab:pincer_performance}
        \begin{tabular}{@{}llrr@{}}
        \toprule
        \textbf{Case Type} & \textbf{Evaluation Level} & \textbf{Number of patients} & \textbf{Rate} \\
        \midrule
        \multicolumn{4}{l}{\textbf{Validated Prescribing Safety Indicator (n=52)}} \\
        \midrule
        & \textit{Level 1: Issue Identification} & & \\
        & \quad Issue identified & 52 & 100.0\% \\
        & \quad Issue not identified & 0 & 0.0\% \\
        \cmidrule(lr){2-4}
        & \textit{Level 2: Issue Correctness} & & \\
        & \quad Correct issue & 43 & 82.7\% \\
        & \quad Incorrect issue & 9 & 17.3\% \\
        \cmidrule(lr){2-4}
        & \textit{Level 3: Intervention Appropriateness} & & \\
        & \quad Correct intervention & 40 & 76.9\% \\
        & \quad Incorrect intervention & 3 & 5.8\% \\
        & \quad Did not reach Level 3 & 9 & 17.3\% \\
        \bottomrule
        \end{tabular}
        \caption{Performance Against Prescribing Safety Indicators. Analysis restricted to 52 cases where clinician validated that the indicator flag correctly identified an issue requiring intervention (from 73 total indicator-positive cases after excluding filters 16 and 43).}
    \end{table}

    Against clinician-validated indicator cases, the System achieved perfect Level 1 sensitivity, identifying all 52 cases requiring intervention (100\%). At Level 2, the system correctly identified the specific indicator in 43 cases (82.7\%), with 9 cases where the System flagged an issue but did not identify the indicator was present. At Level 3, among the 43 correctly identified issues, appropriate interventions were proposed in 40 cases (76.9\% of all positive cases, 93.0\% conditional on Level 2 success).

\subsection{Performance by Filter Type}
    \label{app:prescribing-safety-indicators-performance-by-type}

    Performance varied substantially across the 8 validated indicators. Mean clinician scores ranged from 0.50 (Filter 23: combined hormonal contraception with obesity, BMI $\geq$ 40) to 0.80 (Filter 28: NSAID + peptic ulcer), with most filters clustering between 0.60 and 0.80.

    Filter 28 (NSAID + peptic ulcer) achieved the highest mean score of 0.80 with lower performance from Filter 23 (0.50), Filter 6 (beta-blocker + asthma, 0.60), and Filter 55 (methotrexate without LFT monitoring, 0.59).

    Interpretation is complicated by the small number of clinician-validated cases per filter (range: 6-9 patients per filter after clinician disagreement exclusions), leading to wide confidence intervals that limit robust conclusions about filter-specific performance patterns.

    \begin{figure}[ht]
        \centering
        \includegraphics[width=0.8\textwidth]{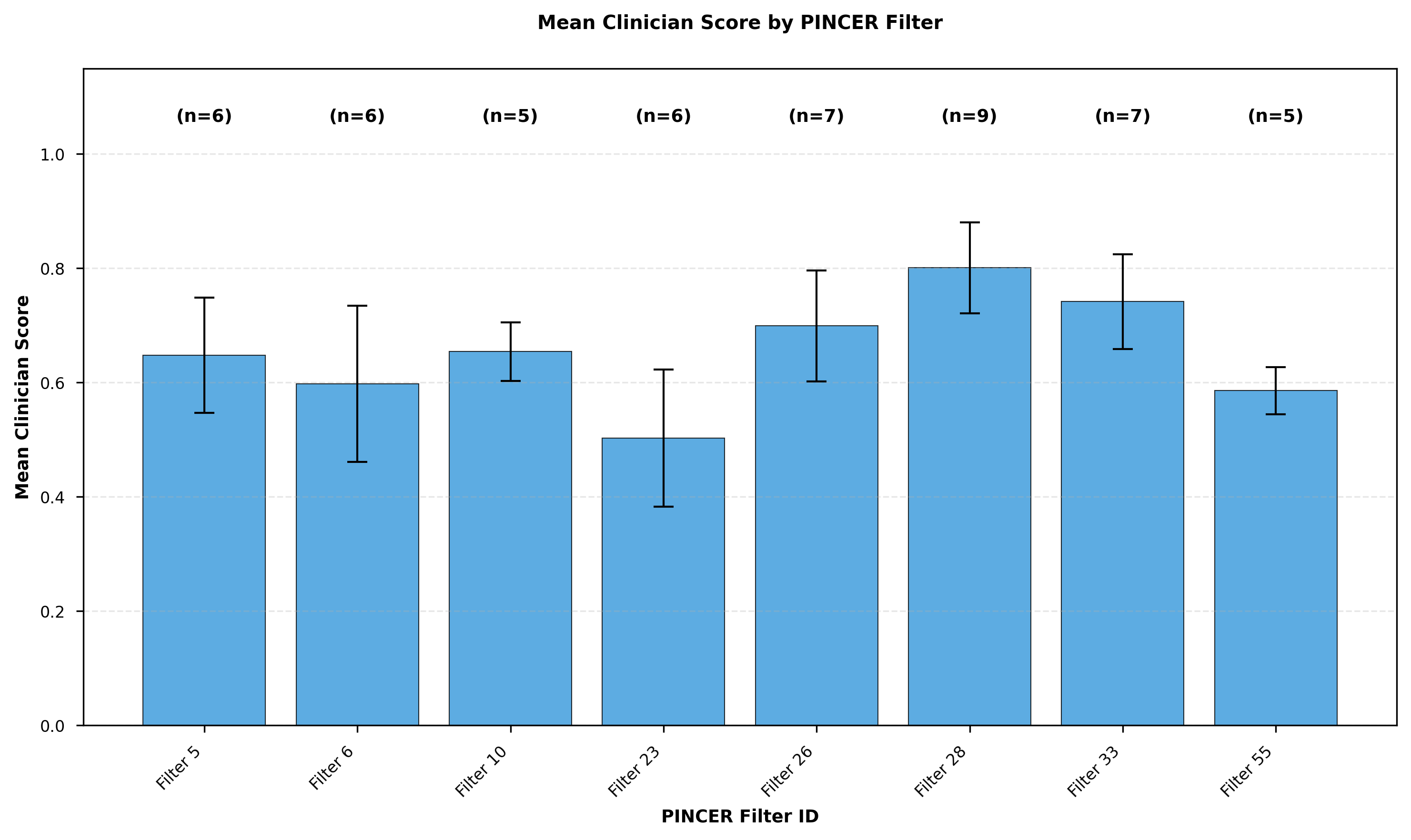}
        \caption{the System's performance by indicator. Mean clinician scores shown with standard error bars. Sample sizes per filter range from 6-9 patients (only cases where clinician validated intervention was needed). Eight filters shown after excluding filters 16 and 43 due to implementation errors.}
        \label{fig:performance-by-filter}
    \end{figure}
\section{Population-Level Performance Assessment}
\label{app:population-performance-assessment}

        The primary evaluation employed stratified sampling (100 indicator-positive + 100 complexity-matched indicator-negative + 100 random indicator-negative) to ensure sufficient representation for detailed failure mode analysis. However, this 50:50 design does not reflect true population distribution, limiting generalisability of raw performance metrics.

        To estimate unbiased population-level performance, we consider the subset of 100 random indicator-negative patients. After data quality filtering, 95 patients remained evaluable.

        The System's flagging rate in this unselected population establishes the base rate for deployment. We measured performance within two strata: System-flagged patients and System-not-flagged patients. To extrapolate to population-level estimates, we calculated stratum-specific performance rates (positive predictive value among flagged; negative predictive value among not-flagged) and re-weighted by the observed population flagging rate. This stratified re-weighting approach yields unbiased population-level sensitivity, specificity, and predictive values that account for the true distribution of System activation in clinical practice.
    
        Among the 51 System-flagged patients, clinician review confirmed clinical issues requiring intervention in 46 cases (90.2\% positive predictive value) with 5 false positives (9.8\%). Among the 44 patients the System did not flag, clinician review found 44 true negatives and 0 false negatives (100\% negative predictive value in this stratum).

        To extrapolate from the evaluated sample to the broader population, we calculated performance rates within each stratum (flagged vs not-flagged) and reweighted by the observed population flagging rate (46.3\% flagged, 53.6\% not-flagged). This stratified reweighting approach accounts for the true distribution of System activation in clinical practice and yields unbiased population-level estimates.

        \begin{table}[h]
            \centering
            \caption{Population-level performance estimates}
            \label{tab:performance}
            \begin{tabular}{lc}
            \toprule
            \textbf{Metric} & \textbf{Value} \\
            \midrule
            Population prevalence & 41.5\% \\
            Sensitivity & 100\% \\
            Specificity & 92.3\% \\
            Positive Predictive Value & 90.2\% \\
            Negative Predictive Value & 100\% \\
            Overall Accuracy & 95.5\% \\
            Cohen's $\kappa$ & 0.909 \\
            F1 Score & 0.948 \\
            \bottomrule
            \end{tabular}
        \end{table}
\section{Ground Truth Establishment Challenges in Structured EHR Data}
\label{app:ground-truth-challenges}
Prior to conducting this evaluation, we investigated whether reliable ground truth could be established from SNOMED codes alone. We found it was not possible to generate a reliable signal, let alone a reliable indicator of the underlying cause for a possible medication change. Medication change rates were nearly identical between Structured Medication Reviews that contained SNOMED codes indicating an issue had been found (e.g. ``Recommendation to stop drug treatment") and those where opposite codes were found (e.g. ``Recommendation to continue with drug treatment") with 30.8\% of medications changing within a three month window versus 30.2\%. Changing this window from 3 months showed that there was no clear temporal threshold where the difference in medication changes was particularly acute.

Even on assessment of the indicators there was disagreement with the clinician in 30.1\% of cases (Table \ref{tab:clinician_indicator_agreement}) due to the nuances of the case. Nevertheless we saw that the use of such filters was the only way to generate a high quality ground truth at scale across the full dataset. Challenges in such an evaluation would remain around negative case identification.

Irrespective of which ground truth you would apply from the structured EHR data, it's not possible to capture nuance around the decision making and this prevents you from making an interesting analysis of the failure modes. To be able to conduct an evaluation at scale without clinician data requires access beyond the structured data to account for patient preference, adherence, and other non-coded information.

\section{Failure Vignettes}
\label{app:failure-vignettes}
{\scriptsize
            \begin{longtable}{p{0.03\textwidth}p{0.19\textwidth}p{0.37\textwidth}p{0.37\textwidth}}
                \label{tab:overconfidence-examples} \\
                \toprule
                \textbf{ID} & \textbf{Failure Reason / Mode} & \textbf{Patient Context} & \textbf{System Intervention and How to Improve} \\
                \midrule
                \endfirsthead
            
                \multicolumn{4}{c}{\tablename\ \thetable{} -- continued from previous page} \\
                \toprule
                \textbf{ID} & \textbf{Failure Reason / Mode} & \textbf{Patient Context} & \textbf{System Intervention and How to Improve} \\
                \midrule
                \endhead
            
                \midrule
                \multicolumn{4}{r}{Continued on next page} \\
                \endfoot
            
                \bottomrule
                \endlastfoot
            
                \multicolumn{4}{l}{\textbf{Failure Reason: Overconfidence}} \\
                \midrule
                \multicolumn{4}{l}{\textit{Premature action without information gathering}} \\
                \midrule
                \hypertarget{vignette-1}{1} & Acting on historical information & 32-year-old woman with documented pregnancy 5 months prior, continuing combined oral contraception & System recommended immediate cessation, but pregnancy status required verification as patient may no longer be pregnant or may have experienced pregnancy loss \\
                \midrule
                \hypertarget{vignette-2}{2} & Making medication changes without understanding reason & 76 year old woman on long-term warfarin for atrial fibrillation, documented heart failure, and recent verapamil prescription (160mg three times daily) & System identified contraindication of verapamil with known heart failure and recommended immediate stop. However, understanding the original reason for verapamil prescription (rate control in AF) before changing it would be essential (consideration should be given to safer alternatives and discussion with specialist) \\
                \midrule
                \hypertarget{vignette-3}{3} & Discontinuing specialist medications without consultation & Patient receiving methotrexate for rheumatological disease alongside potentially interacting naproxen, flagged for NSAID-methotrexate interaction & System recommended holding methotrexate pending alcohol reduction counselling. However, methotrexate is specialist prescribing only and holding it without specialist consultation could precipitate relapse—discussion with rheumatologist essential \\
                \midrule
                \hypertarget{vignette-4}{4} & Non-medication pathways not in coded records & 44-year-old man with moderate-severe depression (PHQ-9 17/27) and severe anxiety (GAD-7 16/21) had sertraline prescriptions listed as not being taken & System recommended starting sertraline 50mg daily to address untreated depression and anxiety. However, the patient may be accessing psychological therapies such as talking therapy or Cognitive behaviour therapy, support from mental health teams, or other non-pharmacological interventions for mood disorder \\
                \midrule
                \hypertarget{vignette-5}{5} & Discontinuing specialist medications without consultation & Patient on carbamazepine for schizophrenia management with identified hyponatraemia and neutropenia & System recommended discontinuing or dose-reducing carbamazepine. However, carbamazepine may have multiple indications including seizure control or mood stabilization, and abrupt discontinuation could cause severe mental health deterioration, requiring discussion with mental health team \\
                \midrule
                \multicolumn{4}{l}{\textit{Overlooked clinical issues}} \\
                \midrule
                \hypertarget{vignette-6}{6} & Missed anticholinergic burden & 67-year-old woman with COPD, correctly flagged for duplicate atorvastatin and propranolol use in obstructive airway disease, had anticholinergic burden score of 6 & Anticholinergic burden not mentioned or addressed by System \\
                \midrule
                \hypertarget{vignette-7}{7} & Missed folic acid supplementation & 70-year-old man on weekly subcutaneous methotrexate (Metoject) for rheumatoid arthritis, correctly identified as having verapamil contraindicated in systolic heart failure & Folic acid alongside methotrexate was needed, as previous folic acid not currently active at medication review. This was not suggested by the System \\
                \midrule
                \hypertarget{vignette-8}{8} & Inadequate electrolyte assessment & 66-year-old woman with persistent hyponatraemia (sodium 123-125 mmol/L) & System suggested no required intervention when investigation of hyponatraemia causes is needed especially in this age group, specifically paired serum and urine osmolalities, early morning cortisol, and thyroid function tests \\
                \midrule
                \multicolumn{4}{l}{\textbf{Failure Reason: Protocol vs Patient gap}} \\
                \midrule
                \multicolumn{4}{l}{\textit{Not considering patient specific context - Palliative care and frailty}} \\
                \midrule
                \hypertarget{vignette-9}{9} & End-of-life care planning & 80-year-old woman with coronary heart disease, heart failure, chronic kidney disease, and documented frailty (frailty index 0.33, 69\% mortality risk) had discontinued aspirin, statin, ACE inhibitor, and beta-blocker 9 months prior & System flagged all these omissions and recommended re-initiation of all of them. While these may be correct for someone younger and physiologically robust, in this case their withdrawal represented appropriate end-of-life care planning \\
                \midrule
                \hypertarget{vignette-10}{10} & Palliative care priorities & 90-year-old woman with dementia, atrial fibrillation, hypertension, osteoporosis and secondary hyperparathyroidism on the palliative care register had missing calcium and vitamin D supplementation & System recommended starting calcium and vitamin D—less concerning for a patient with such advanced disease and palliation \\
                \midrule
                \hypertarget{vignette-11}{11} & Age-appropriate targets & 82-year-old woman with severe hypertension (systolic 192 mmHg) on amlodipine monotherapy & System flagged for uncontrolled blood pressure and recommended initiating ramipril 5mg once daily and targeting blood pressure below 140/90 mmHg. However, frailty and falls risk may outweigh the benefit of adding an additional agent, and blood pressure targets could be more relaxed in patients over 80 \\
                \midrule
                \multicolumn{4}{l}{\textit{Not considering patient specific context - Complex treatment regimens}} \\
                \midrule
                \hypertarget{vignette-12}{12} & Competing medication indications & Patient with atrial fibrillation, heart failure, and documented hypotension on multiple antihypertensive agents including nebivolol (beta-blocker), amlodipine, furosemide, and spironolactone & System correctly identified hypotension warranted reducing antihypertensive burden and recommended discontinuing nebivolol, the lowest-dose beta-blocker. However, discontinuing the beta-blocker risked precipitating fast atrial fibrillation and hospitalization; amlodipine would be safer choice to withdraw. System's recommendation failed to acknowledge this dual indication or signal uncertainty \\
                \midrule
                \hypertarget{vignette-13}{13} & Nuanced risk-benefit balance & Patient with severe mental illness on flupentixol prescribed procyclidine prophylactically to prevent extrapyramidal symptoms & System flagged procyclidine as potentially inappropriate anticholinergic therapy and recommended discontinuation. However, procyclidine was used prophylactically because some features of extrapyramidal syndrome, particularly tardive dyskinesia, may be irreversible if allowed to develop, requiring shared decision-making about competing risks \\
                \midrule
                \multicolumn{4}{l}{\textit{Rigid application of safety thresholds}} \\
                \midrule
                \hypertarget{vignette-14}{14} & Rigid threshold application & 80-year-old non-diabetic patient with blood pressure readings of 130-140/60-80 mmHg & System flagged as untreated hypertension requiring intervention when this range was acceptable for an 80-year-old non-diabetic patient \\
                \midrule
                \hypertarget{vignette-15}{15} & Misapplied dosing criteria & 68-year-old male with creatine 135~$\mu$mol/L on 5mg apixaban twice daily & System recommended dose reduction based on "apixaban dose exceeding renal dosing recommendation"  when guidelines require additional criteria to be met (age$>$80 or body weight $<$60kg) \\
                \midrule
                \hypertarget{vignette-16}{16} & Overstated interaction risk & Patient on low-dose atorvastatin-amlodipine & System flagged for myopathy risk when the interaction is only significant with simvastatin or high-dose atorvastatin \\
                \midrule
                \hypertarget{vignette-17}{17} & Inappropriate frailty assessment & 41-year-old woman with history of depression and anxiety on propranolol with frailty codes for falls and dizziness & System recommended discontinuation of propranolol based on frailty codes when there were no further documented concerns in the record and multiple entries noting anxiety \\
                \midrule
                \multicolumn{4}{l}{\textit{Missed deprescription opportunities}} \\
                \midrule
                \hypertarget{vignette-18}{18} & Palliative care medication burden & 87-year-old man on palliative care register with dementia and documented falls taking simvastatin and aspirin 75mg daily & System correctly identified duplicate bisoprolol prescriptions and donepezil-bisoprolol interaction contributing to falls, but missed opportunities to question ongoing value of statin in someone on palliative care register with dementia, and to discuss whether aspirin risk was outweighing its benefit \\
                \midrule
                \hypertarget{vignette-19}{19} & Palliative care medication burden & 84-year-old man with vascular dementia on palliative care register taking warfarin, clopidogrel, and atorvastatin 80mg & System recommended stopping clopidogrel to reduce bleeding risk and reducing atorvastatin to 20mg for elevated liver enzymes. While stopping clopidogrel was appropriate, the statin could be stopped altogether in palliative care if the patient agreed, as risk outweighed benefit \\
                \midrule
                \hypertarget{vignette-20}{20} & Medication without clear indication & 75-year-old woman with rheumatoid arthritis on methotrexate taking omeprazole 20mg daily & System correctly flagged for missing folic acid supplementation, but there was no obvious indication for the omeprazole 20mg daily, which should have been flagged as potentially unnecessary \\
                \midrule
                \multicolumn{4}{l}{\textbf{Failure Reason: Protocol vs Practice gap}} \\
                \midrule
                \multicolumn{4}{l}{\textit{Duplicate Prescription Errors}} \\
                \midrule
                \hypertarget{vignette-21}{21} & Multiple tablet strengths to achieve target dose & 56-year-old woman with multiple comorbidities prescribed ramipril 2.5mg and ramipril 1.25mg concurrently & System flagged as duplicate dosing creating excess exposure and recommended stopping the 1.25mg tablet. However, the intended total daily dose was 3.75mg—which cannot be achieved with a single tablet strength \\
                \midrule
                \hypertarget{vignette-22}{22} & Multiple tablet strengths to achieve target dose & 56-year-old woman with essential hypertension prescribed perindopril 2mg and perindopril 4mg tablets, both taken each morning, achieving a total daily dose of 6mg & System flagged as duplicate prescribing, and recommended consolidating to a single 6mg tablet. However, perindopril 6mg tablets do not exist—the intended total daily dose of 6mg requires using two tablet strengths concurrently \\
                \midrule
                \hypertarget{vignette-23}{23} & Tapering or transition phases misinterpreted as concurrent duplicates & 30-year-old pregnant woman had two venlafaxine prescriptions listed (Vensir XL 225mg once daily and 150mg twice daily) & System calculated a total dose of 525mg daily and recommended discontinuing both prescriptions as duplicate high-dose prescribing. However, this was part of a dose-optimisation strategy—the 225mg dose had replaced the 150mg dose rather than added to it, and the record noted that the prescriptions were not being taken concurrently \\
                \midrule
                \multicolumn{4}{l}{\textit{Healthcare system context}} \\
                \midrule
                \hypertarget{vignette-24}{24} & Prescription records vs actual patient exposure & 84-year-old woman on warfarin prescribed flucloxacillin with explicit note to stop simvastatin during the antibiotic course & System flagged simvastatin as still active and recommended immediate discontinuation. However, the patient was clearly advised to discontinue simvastatin during antibiotic therapy—the prescription note indicated the medication was not actually being taken \\
                \midrule
                \hypertarget{vignette-25}{25} & Prescription records vs actual patient exposure & 95-year-old man had two ACE inhibitors listed as active prescriptions: ramipril 1.25mg daily and quinapril 5mg daily & System flagged duplicate therapy carrying high risk of hypotension and acute kidney injury. However, a free text annotation stated ramipril was only being prescribed until quinapril came back in stock due to supply shortage—the patient was never actually receiving both ACE inhibitors simultaneously \\
                \midrule
                \hypertarget{vignette-26}{26} & Medications managed outside primary care record & 39-year-old woman with type 1 diabetes had no insulin prescription recorded despite markedly elevated HbA1c (65-70 mmol/mol) and ongoing glucose monitoring with FreeStyle Libre sensor & System recommended initiating insulin therapy urgently to prevent diabetic ketoacidosis. However, the patient must be receiving insulin as a type 1 diabetic; her insulin supply comes from secondary care, likely via continuous insulin pump, which is not captured in the primary care medication record \\
                \midrule
                \multicolumn{4}{l}{\textbf{Failure Reason: Coherent but factually incorrect}} \\
                \midrule
                \multicolumn{4}{l}{\textit{Hallucinations}} \\
                \midrule
                \hypertarget{vignette-27}{27} & Misidentified drug composition - clopidogrel & Patient prescribed Monomil XL, a brand of isosorbide mononitrate & System indicated that it contained clopidogrel and identified dual therapy. However, Monomil XL does not contain clopidogrel \\
                \midrule
                \hypertarget{vignette-28}{28} & Misidentified drug composition - calcium channel blocker & Patient prescribed Monomil XL, a brand of isosorbide mononitrate & System indicated that it was a calcium channel blocker and identified dual therapy. However, Monomil XL is not a calcium channel blocker \\
                \midrule
                \hypertarget{vignette-29}{29} & Misidentified drug composition - opioid & Patient prescribed Monomil XL, a brand of isosorbide mononitrate & System indicated that it was an opioid. However, Monomil XL is not an opioid \\
                \midrule
                \hypertarget{vignette-30}{30} & Misidentified drug composition - tramadol and acetaminophen & Patient prescribed Zapain & System incorrectly thought it contained tramadol and acetaminophen and made recommendations accordingly, whereas Zapain in fact contains a combination of codeine and paracetamol \\
                \midrule
                \hypertarget{vignette-31}{31} & Incorrect VTE risk with transdermal oestrogen & Patient prescribed estradiol 0.06\% transdermal gel for menopausal symptoms, current smoker with prior pulmonary embolism & System noted that the transdermal oestrogen might increase the risk of venous thromboembolism, referencing the fact that the patient was a current smoker with prior pulmonary embolism. Transdermal oestrogens are not associated with an increased risk of venous thromboembolism \\
                \midrule
                \multicolumn{4}{l}{\textit{Pharmacological knowledge gaps}} \\
                \midrule
                \hypertarget{vignette-32}{32} & Topical NSAID systemic risk overestimation & 82-year-old woman with chronic kidney disease stage 3 (eGFR 50 mL/min) on furosemide and candesartan using Fenbid 5 percent gel for musculoskeletal pain & System flagged topical NSAID use in chronic kidney disease on loop diuretic and ARB as requiring intervention and recommended stopping immediately due to risk of acute kidney injury. However, systemic absorption of NSAID gels is much below the threshold where systemic side effects might be expected \\
                \midrule
                \hypertarget{vignette-33}{33} & Topical NSAID systemic risk overestimation & 87-year-old woman with peptic ulcer disease using Fenbid 10 percent gel & System flagged NSAID use despite having a current peptic ulcer—the System also flagged oral naproxen in this patient, but failed to distinguish which posed the actual risk. However, systemic absorption of NSAID gels is much below the threshold where systemic side effects might be expected \\
                \midrule
                \hypertarget{vignette-34}{34} & Illogical intervention sequence & 54-year-old man with peptic ulcer history taking ibuprofen 200mg three times daily & System correctly flagged for NSAID use in ulcer-prone patient. The System recommended discontinuing ibuprofen immediately, prescribing paracetamol for analgesia, and simultaneously starting omeprazole 20mg daily for ulcer protection. However, stopping the ibuprofen and starting a gastro-protective agent at the same time is illogical—stopping the NSAID removes the risk factor so PPI is not required \\
                \midrule
                \hypertarget{vignette-35}{35} & Misidentified therapeutic purpose & 85-year-old man with falls and hypotension taking losartan, amlodipine, and tamsulosin & System recommended discontinuing tamsulosin to reduce orthostatic hypotension from "triple anti-hypertensive therapy." However, tamsulosin is used for prostate-related urinary symptoms, not blood pressure control—any review should focus on urinary symptom management and whether safer alternatives exist for patients with falls risk \\
                \midrule
                \multicolumn{4}{l}{\textit{Guidelines misapplication}} \\
                \midrule
                \hypertarget{vignette-36}{36} & Incorrect statin threshold & 58-year-old woman with ischaemic heart disease and a QRISK2 10-year CVD risk of 7.8\% with persistently elevated blood pressure but not on statin therapy & System recommended initiating atorvastatin 20mg daily for established cardiovascular disease, although without referencing guidelines. However, the threshold for statin therapy is greater than 10\% QRISK over 10 years, above the patient's current level \\
                \midrule
                \hypertarget{vignette-37}{37} & Misunderstood BP measurement guidelines & 67-year-old man with clinic blood pressure of 140/90 mmHg recorded 1 year 7 months prior but home readings of 124/76 mmHg & System flagged uncontrolled hypertension with no antihypertensive therapy and recommended starting ramipril 2.5mg daily. However, UK guidelines state home readings are more accurate than clinic measurements, and treatment would not be recommended on the basis of these readings \\
                \midrule
                \hypertarget{vignette-38}{38} & Incorrect maximum dose & 49-year-old man taking citalopram 30mg daily (1.5 tablets of 20mg) & System flagged for exceeding the age-appropriate maximum dose, stating ``Citalopram dose exceeds age-appropriate maximum (30mg daily $>$ 20mg recommended for $>$45yr)". However, the maximum dose is 40mg daily, making the current 30mg dose appropriate \\
                \midrule
                \hypertarget{vignette-39}{39} & Intervention not supported by guidelines & 54-year-old man with diabetes who experienced a first unprovoked epileptic seizure 9 months prior with no antiepileptic medication prescribed & System flagged this as an issue requiring intervention. However, epilepsy guidelines and ILAE guidelines advise treatment is not necessary after first seizure unless considered high risk (greater than 60 percent) of recurrence according to NICE NG217, and the seizure was possibly due to low blood sugar which would not be classified as epileptiform \\
                \midrule
                \multicolumn{4}{l}{\textbf{Failure Reason: Process blindness}} \\
                \midrule
                \hypertarget{vignette-40}{40} & Unsafe medication transitions without risk assessment & 83-year-old frail patient with atrial fibrillation and no current anticoagulation & System advised to start apixaban. However, this was too aggressive without first calculating bleeding risk (HAS-BLED score) and making a shared decision with the patient about risks versus benefits \\
                \midrule
                \hypertarget{vignette-41}{41} & Unsafe medication transitions without confirming diagnosis & Patient with elevated clinic blood pressure readings & System instructed to start ramipril, when NICE guidelines require home or ambulatory monitoring to confirm the diagnosis of hypertension before initiating treatment \\
                \midrule
                \hypertarget{vignette-42}{42} & Immediate discontinuation requiring gradual tapering & 76-year-old woman with cognitive impairment and constipation taking amitriptyline 10mg nightly & System identified the high anticholinergic burden and recommended stopping amitriptyline immediately and initiating a safer alternative such as sertraline. However, while ultimately discontinuation of amitriptyline was appropriate, it requires tapering over several weeks \\
                \midrule
                \hypertarget{vignette-43}{43} & Immediate discontinuation requiring gradual tapering & 54-year-old woman on both sertraline 100mg daily and duloxetine 60mg nightly & System flagged for serotonin syndrome risk and suggested immediate discontinuation of duloxetine—tapering would be required instead \\
                \midrule
                \hypertarget{vignette-44}{44} & Discontinuing therapy without ensuring alternative management & 27-year-old woman with severe obesity (BMI 46.5 kg/m²) taking combined oral contraceptive (Rigevidon) & System identified the reduced effectiveness due to high BMI and recommended discontinuing Rigevidon then subsequently discussing alternative contraceptive methods. However, stopping the contraceptive pill before organizing an alternative would increase pregnancy risk rather than reduce it—a discussion was necessary before any changes to the patient's contraception \\
                \midrule
                \hypertarget{vignette-45}{45} & Discontinuing therapy without ensuring alternative management & Patient with type 2 diabetes, advanced CKD (eGFR 30 mL/min), and poor glycemic control (HbA1c 77 mmol/mol) on metformin 500mg three times daily & System recommended stopping metformin without suggesting an alternative glucose-lowering agent. However, simply discontinuing metformin would worsen the already poor glycemic control and accelerate renal decline—an alternative agent was necessary \\
                \midrule
            \end{longtable}
            }

\section{Scoring Methodology}
\label{app:scoring}

This appendix provides full specification of the scoring methodologies used for clinician evaluation (Section~\ref{sec:scoring-clinician}) and automated evaluation (Section~\ref{sec:scoring-automated}).

\subsection{Clinician Composite Score}
\label{sec:scoring-clinician}

The clinician evaluated each case across multiple dimensions but did not provide an overall score. To enable quantitative analysis, we calculated a composite score ($S_{\text{clinician}}$, range 0--1) combining issue identification accuracy and intervention appropriateness.

\subsubsection{Issue Identification}

Issue identification was assessed using precision and recall:

\begin{equation}
    P_{\text{issue}} = \frac{|\text{correct issues identified by System}|}{|\text{total issues identified by System}|}
\end{equation}

Recall was assessed categorically based on clinician judgment of missed issues:

\begin{equation}
    R_{\text{issue}} = \begin{cases}
        1.0 & \text{no missed issues} \\
        0.5 & \text{missed issues, but some correct issues identified} \\
        0.0 & \text{missed issues, no correct issues identified}
    \end{cases}
\end{equation}

These were combined into an $F_1$ score:

\begin{equation}
    F_{1,\text{issue}} = \frac{2 \cdot P_{\text{issue}} \cdot R_{\text{issue}}}{P_{\text{issue}} + R_{\text{issue}}}
\end{equation}

\subsubsection{Intervention Appropriateness}

Intervention appropriateness was scored categorically:

\begin{equation}
    S_{\text{intervention}} = \begin{cases}
        1.0 & \text{intervention fully correct} \\
        0.5 & \text{intervention partially correct} \\
        0.0 & \text{intervention incorrect or missing when needed}
    \end{cases}
\end{equation}

\subsubsection{Composite Score}

The final clinician composite score averages issue identification and intervention appropriateness:

\begin{equation}
    S_{\text{clinician}} = \begin{cases}
        \frac{1}{2}(F_{1,\text{issue}} + S_{\text{intervention}}) & \text{if } F_{1,\text{issue}} \text{ is defined (i.e., issues were identified)} \\[0.5em]
        S_{\text{intervention}} & \text{otherwise (true negative cases)}
    \end{cases}
\end{equation}

The partial credit value (0.5) for both recall and intervention appropriateness reflects clinical judgment that partially correct outputs retain meaningful utility---a recommendation that identifies some but not all issues, or proposes a reasonable but suboptimal intervention, is more valuable than a completely incorrect assessment while still falling short of full clinical acceptability.

\subsection{Automated Scorer}
\label{sec:scoring-automated}

For scalable evaluation across multiple models and repeated runs, we developed an automated scorer using clinician feedback as ground truth.

\subsubsection{Ground Truth Establishment}

Clinician feedback on the 277 evaluated patients was synthesised into structured ground truth using \texttt{gpt-oss-120b}. For each patient, this produced:
\begin{itemize}
    \item A validated list of clinical issues requiring intervention (or confirmation that no issues existed)
    \item A validated list of appropriate interventions
\end{itemize}

\subsubsection{Automated Scoring Process}

For each System evaluation, a separate LLM judge assessed alignment with the ground truth. The judge determined:
\begin{itemize}
    \item Which System-identified issues matched ground truth issues
    \item Which ground truth issues were missed by the System
    \item Which System-proposed interventions matched ground truth interventions
\end{itemize}

\subsubsection{Score Calculation}

Precision and recall were calculated for both issues and interventions:

\begin{align}
    P_{\text{issue}} &= \frac{|\text{System issues matching ground truth}|}{|\text{total System issues}|}\\[0.5em]
    R_{\text{issue}} &= \frac{|\text{ground truth issues matched by System}|}{|\text{total ground truth issues}|}\\[0.5em]
    P_{\text{intervention}} &= \frac{|\text{System interventions matching ground truth}|}{|\text{total System interventions}|}\\[0.5em]
    R_{\text{intervention}} &= \frac{|\text{ground truth interventions matched by System}|}{|\text{total ground truth interventions}|}
\end{align}

These were combined into $F_1$ scores:

\begin{align}
    F_{1,\text{issue}} &= \frac{2 \cdot P_{\text{issue}} \cdot R_{\text{issue}}}{P_{\text{issue}} + R_{\text{issue}}}\\[0.5em]
    F_{1,\text{intervention}} &= \frac{2 \cdot P_{\text{intervention}} \cdot R_{\text{intervention}}}{P_{\text{intervention}} + R_{\text{intervention}}}
\end{align}

The final automated score handles three cases:

\begin{equation}
    S_{\text{automated}} = \begin{cases}
        1.0 & \text{true negative} \\[0.5em]
        0.0 & \text{false positive or false negative} \\[0.5em]
        \frac{1}{2}(F_{1,\text{issue}} + F_{1,\text{intervention}}) & \text{both flagged issues (evaluate quality)}
    \end{cases}
\end{equation}

\subsubsection{Design Considerations}

The automated scorer was designed for research use to enable comparison across models and configurations. Several design choices reflect this context:

\textbf{Equal weighting of false positives and false negatives.} The scorer assigns 0.0 to both false positives (System flags issue, ground truth has none) and false negatives (System misses issue in ground truth). For deployment contexts, asymmetric weighting could prioritise sensitivity, reflecting the greater harm from missing genuine medication safety risks compared to flagging cases that prove acceptable on review.

\textbf{LLM-as-judge limitations.} Using an LLM to assess alignment between System outputs and ground truth introduces potential for systematic bias, particularly for edge cases where clinical judgment is required. The scorer was validated against a subset of manually-scored cases to ensure reasonable alignment, but should not be interpreted as equivalent to expert clinician evaluation.

\textbf{Ground truth constraints.} The ground truth reflects a single expert clinician's assessment. Inter-rater reliability was not formally assessed due to resource constraints within the Trusted Research Environment. The 30.1\% disagreement rate between clinician assessment and validated prescribing safety indicators (Appendix~\ref{app:prescribing-safety-indicators}) suggests meaningful variability exists even among expert judgments.
\section{Technical Infrastructure and Data Processing}
\label{app:technical-infrastructure}
\subsection{Computing Infrastructure}

    The Trusted Research Environment (TRE) included two Azure virtual machines with a shared filesystem. The CPU node was an Azure Standard E48as v6 instance. A secondary GPU node with two NVIDIA H100 GPUs enabled local hosting of large language models (LLMs) with vLLM. This setup enabled processing of the full dataset (2,125,549 patients) within the TRE while also enabling use of leading open source LLMs.
    
\subsection{Data Governance and Ethical Approval}

    Data access and analysis were conducted under the governance framework of the Cheshire \& Merseyside Data Into Action (DIA) programme. The study received approval from the Data Access \& Asset Group (DAAG) following completion of a Data Access Request Form (DARF) and Data Protection Impact Assessment (DPIA). Access was granted for the period June 8, 2025 to November 8, 2025, after which all access was revoked. The study used pseudonymised patient profiles and did not require individual patient consent under the approved data governance arrangements. Data controllers across NHS Cheshire \& Merseyside were informed of the study and retained the right to opt out their patient populations.

\subsection{Data Preprocessing}
    The TRE provided access to historical patient profiles via an export of parquet files. Substantial data preprocessing was required to construct longitudinal patient profiles from multiple linked tables. The preprocessing pipeline is documented in this \href{https://github.com/i-dot-ai/medguard-preprocessing/tree/paper-analysis}{repository}. The preprocessing generated structured patient profiles as Pydantic models containing all available information for a patient. These were then transferred to the GPU for analysis.

    These structured patient profiles had methods to convert them to markdown containing chronologically ordered clinical events: demographics, diagnoses (SNOMED-coded with descriptions), medications (with start/end dates and whether they were active or past prescriptions at the time of the review), laboratory results, hospital episodes, and relevant clinical observations (including height, weight, blood pressure).

    For each patient, the System generated structured outputs validated against Pydantic schemas using default sampling parameters.

    \begin{table}[ht]
        \centering
        \begin{tabular}{@{}llp{7cm}@{}}
        \toprule
        \textbf{Field} & \textbf{Type} & \textbf{Description} \\
        \midrule
        Intervention required & Boolean & Binary flag indicating whether any action needed \\
        Intervention probability & Float (0-1) & Model confidence in intervention necessity \\
        Clinical issues & List & Identified medication safety concerns \\
        Evidence & List & Supporting clinical data for each issue \\
        Proposed interventions & String & Recommended intervention for patient \\
        \bottomrule
        \end{tabular}
        \caption{Structured output schema for System evaluations}
    \end{table}
\section{Clinician Evaluation App}
\label{app:evaluation-app}
To enable systematic clinician evaluation within the Trusted Research Environment, we developed a custom evaluation application using FastAPI (backend) and Next.js (frontend). The application ran locally within the TRE to ensure patient data never left the secure environment.

\begin{figure}[ht]
    \centering
    \begin{subfigure}[t]{0.48\textwidth}
        \centering
        \includegraphics[width=\textwidth]{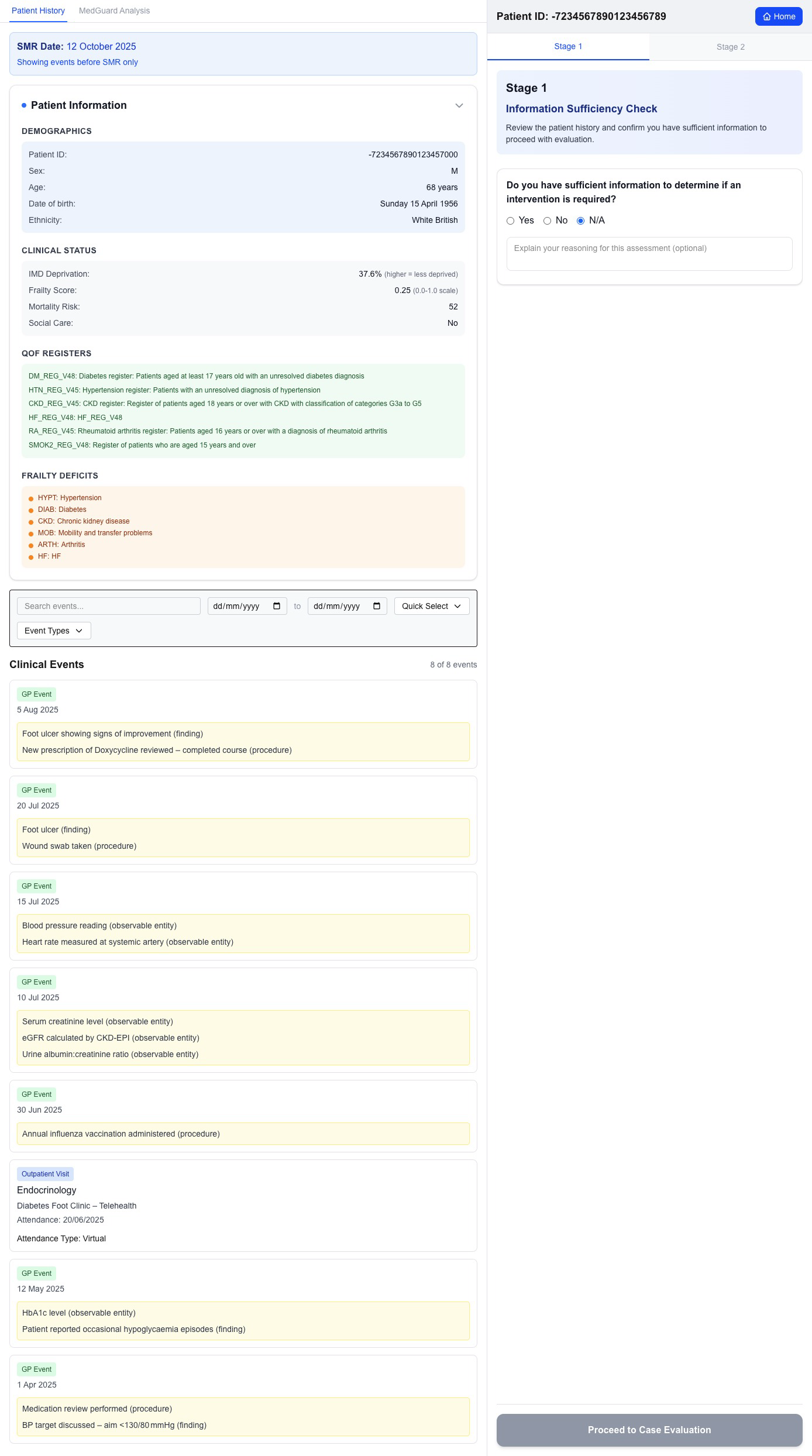}
        \caption{Screen 1: The clinician can look at the structured patient data and decides whether they have sufficient information to determine if an intervention is required for this patient.}
        \label{fig:stage_1_evaluation_app}
    \end{subfigure}
    \hfill
    \begin{subfigure}[t]{0.48\textwidth}
        \centering
        \includegraphics[width=\textwidth]{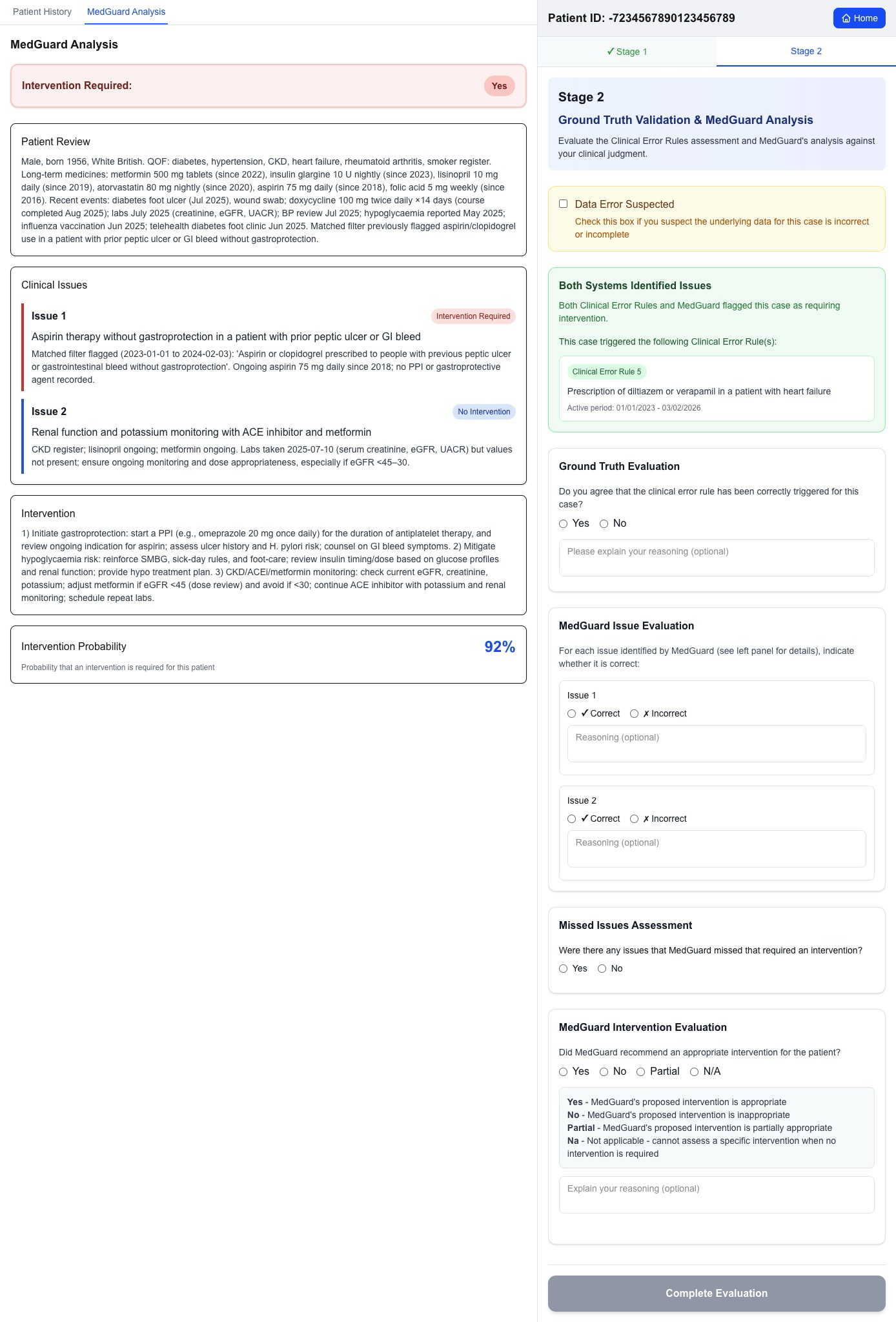}
        \caption{Screen 2: The clinician can look at the structured patient data, and the System's analysis. They then grade the System's issues and interventions.}
        \label{fig:stage_2_evaluation_app}
    \end{subfigure}
    \caption{Clinician Evaluation App}
    \label{fig:evaluation_apps}
\end{figure}

The interface was designed to support the hierarchical evaluation framework. In the first screen (Figure~\ref{fig:evaluation_apps}a), the clinician reviewed the structured patient data---seeing exactly what the System saw---and determined whether sufficient information existed to assess intervention necessity. This stage identified 23 cases excluded for data quality issues. On the second screen (Figure~\ref{fig:evaluation_apps}b), the clinician reviewed both patient data and System analysis side-by-side, grading each identified issue as correct or incorrect, noting any missed issues, and assessing intervention appropriateness.

Results were saved locally to JSON files for subsequent analysis.

\end{appendices}

\end{document}